\documentclass[10pt,twocolumn,letterpaper]{article}

\usepackage[pagenumbers]{cvpr} 

\usepackage{graphicx,tabulary,amsmath,amssymb,booktabs,multirow,multicol,xspace,xcolor,bbding,comment,algorithm,listings,pifont,colortbl,makecell}
\usepackage[pagebackref,breaklinks,colorlinks]{hyperref}

\usepackage[capitalize]{cleveref}
\crefname{section}{Sec.}{Secs.}
\Crefname{section}{Section}{Sections}
\Crefname{table}{Table}{Tables}
\crefname{table}{Tab.}{Tabs.}

\def\x{$\times$}
\def\fsrc{$f_s$\xspace}
\def\ftgt{$f_t$\xspace}
\def\nsrc{$n_s$\xspace}
\def\ntgt{$n_t$\xspace}

\def\vece{\mathbf{e}}
\def\vecv{\mathbf{v}}
\def\vecz{\mathbf{z}}
\def\vecx{\mathbf{x}}
\def\vecf{\mathbf{f}}
\def\vecS{\mathbf{S}}

\def\dd{\mathrm{d}}
\def\pr{\mathbb{P}}
\def\ee{\mathbb{E}}
\def\var{\mathbb{V}}
\def\varm{v}
\def\tr{\mathrm{tr}}

\def\src{source\xspace}
\def\tgt{target\xspace}

\def\mcrop{MultiCrop\xspace}
\def\smix{ScaleMix\xspace}
\def\aaug{AsymAug\xspace}
\def\saug{StrongerAug\xspace}
\def\waug{WeakerAug\xspace}
\def\abn{AsymBN\xspace}
\def\syncbn{SyncBN\xspace}
\def\menc{MeanEnc\xspace}

\definecolor{mygreen}{HTML}{39b54a}  
\definecolor{myred}{HTML}{ea4335}  
\definecolor{verylightgray}{gray}{0.9}

\newcommand{\better}[1]{\textcolor{mygreen}{#1}}
\newcommand{\worse}[1]{\textcolor{myred}{#1}}
\newcommand{\deemph}[1]{\textcolor{gray}{#1}}

\newcommand{\weaker}{$\downarrow$}
\newcommand{\stronger}{$\uparrow$} 
\newcommand{\app}{\raise.17ex\hbox{$\scriptstyle\sim$}}

\newcommand\blfootnote[1]{%
  \begingroup
  \renewcommand\thefootnote{}\footnote{#1}%
  \addtocounter{footnote}{-1}%
  \endgroup
}
\newlength\savewidth\newcommand\shline{\noalign{\global\savewidth\arrayrulewidth
  \global\arrayrulewidth 1pt}\hline\noalign{\global\arrayrulewidth\savewidth}}
\newcommand{\tablestyle}[2]{\setlength{\tabcolsep}{#1}\renewcommand{\arraystretch}{#2}\centering\footnotesize}
\makeatletter\renewcommand\paragraph{\@startsection{paragraph}{4}{\z@}
  {.5em \@plus1ex \@minus.2ex}{-.5em}{\normalfont\normalsize\bfseries}}\makeatother

\newcolumntype{x}[1]{>{\centering\arraybackslash}p{#1pt}}
\newcolumntype{a}[1]{>{\columncolor{verylightgray}\centering\arraybackslash}p{#1pt}}
\newcolumntype{y}[1]{>{\raggedright\arraybackslash}p{#1pt}}
\newcolumntype{z}[1]{>{\raggedleft\arraybackslash}p{#1pt}}

\def\cvprPaperID{}
\def\confName{CVPR}
\def\confYear{2022}

\begin{document}

\title{\vspace{-1em}\Large On the Importance of Asymmetry for Siamese Representation Learning\vspace{-1em}}

\author{Xiao Wang$^{*,\dagger}$ \quad Haoqi Fan$^{1,\dagger}$ \quad Yuandong Tian$^1$ \quad Daisuke Kihara$^2$ \quad Xinlei Chen$^1$\\[1mm]
$^1$Facebook AI Research (FAIR) \qquad $^2$Purdue University\\
\small{Code: \url{https://github.com/facebookresearch/asym-siam}}}

\maketitle

\blfootnote{$*$: work done during internship at FAIR. $\dagger$: equal contribution.}

\begin{abstract}
    Many recent self-supervised frameworks for visual representation learning are based on certain forms of Siamese networks.
    Such networks are conceptually symmetric with two parallel encoders, but often practically asymmetric as numerous mechanisms are devised to break the symmetry. 
    In this work, we conduct a formal study on the importance of asymmetry by explicitly distinguishing the two encoders within the network -- one produces source encodings and the other targets. Our key insight is keeping a relatively lower variance in target than source generally benefits learning. This is empirically justified by our results from five case studies covering different variance-oriented designs, and is aligned with our preliminary theoretical analysis on the baseline. Moreover, we find the improvements from asymmetric designs generalize well to longer training schedules, multiple other frameworks and newer backbones. Finally, the combined effect of several asymmetric designs achieves a state-of-the-art accuracy on ImageNet linear probing and competitive results on downstream transfer.
    We hope our exploration will inspire more research in exploiting asymmetry for Siamese representation learning.
\end{abstract}

\section{Introduction\label{sec:intro}}

Despite different motivations and formulations, many recent un-/self-supervised methods for visual representation learning~\cite{he2020momentum,chen2020simple,grill2020bootstrap,caron2020unsupervised,zbontar2021barlow,caron2021emerging,bardes2021vicreg} are based on certain forms of Siamese networks~\cite{Bromley1994}. Siamese networks are inherently \emph{symmetric}, as the two encoders within such networks share many aspects in design. For example, their model architectures (\eg, ResNet~\cite{he2016deep}) are usually the same; their network weights are often copied over; their input distributions -- typically compositions of multiple data augmentations~\cite{chen2020simple} -- are by default identical; and their outputs are encouraged to be similar for the same image. Such a symmetric structure not only enables straightforward adaptation from off-the-shelf, supervised learning architectures to self-supervised learning, but also introduces a minimal inductive bias to learn representations invariant \wrt various transformations in computer vision~\cite{chen2020exploring}.

However, symmetry is not the only theme in these frameworks. In fact, numerous mechanisms were proposed to break the conceptual symmetry. For example, BYOL~\cite{grill2020bootstrap} and SimSiam~\cite{chen2020exploring} place a special predictor head on one of the encoders, so architecture-wise they are no longer symmetric; MoCo~\cite{he2020momentum} introduces momentum encoder, in which the weights are computed with moving-averages instead of directly copied; SwAV~\cite{caron2020unsupervised} and DINO~\cite{caron2021emerging} additionally adopt a multi-crop~\cite{misra2020self} strategy to enhance the augmentation on one side, shifting the data distribution asymmetric between encoders; even the InfoNCE loss~\cite{oord2018representation} treats outputs from two encoders differently -- one is positive-only and the other also involves negatives. Among them, some specific asymmetric designs are crucial and well-studied (\eg, stop-gradient to prevent collapse~\cite{chen2020exploring}), but the general role of \emph{asymmetry} for Siamese representation learning is yet to be better understood.

\begin{figure}[t]
  \centering
  \includegraphics[width=.7\linewidth]{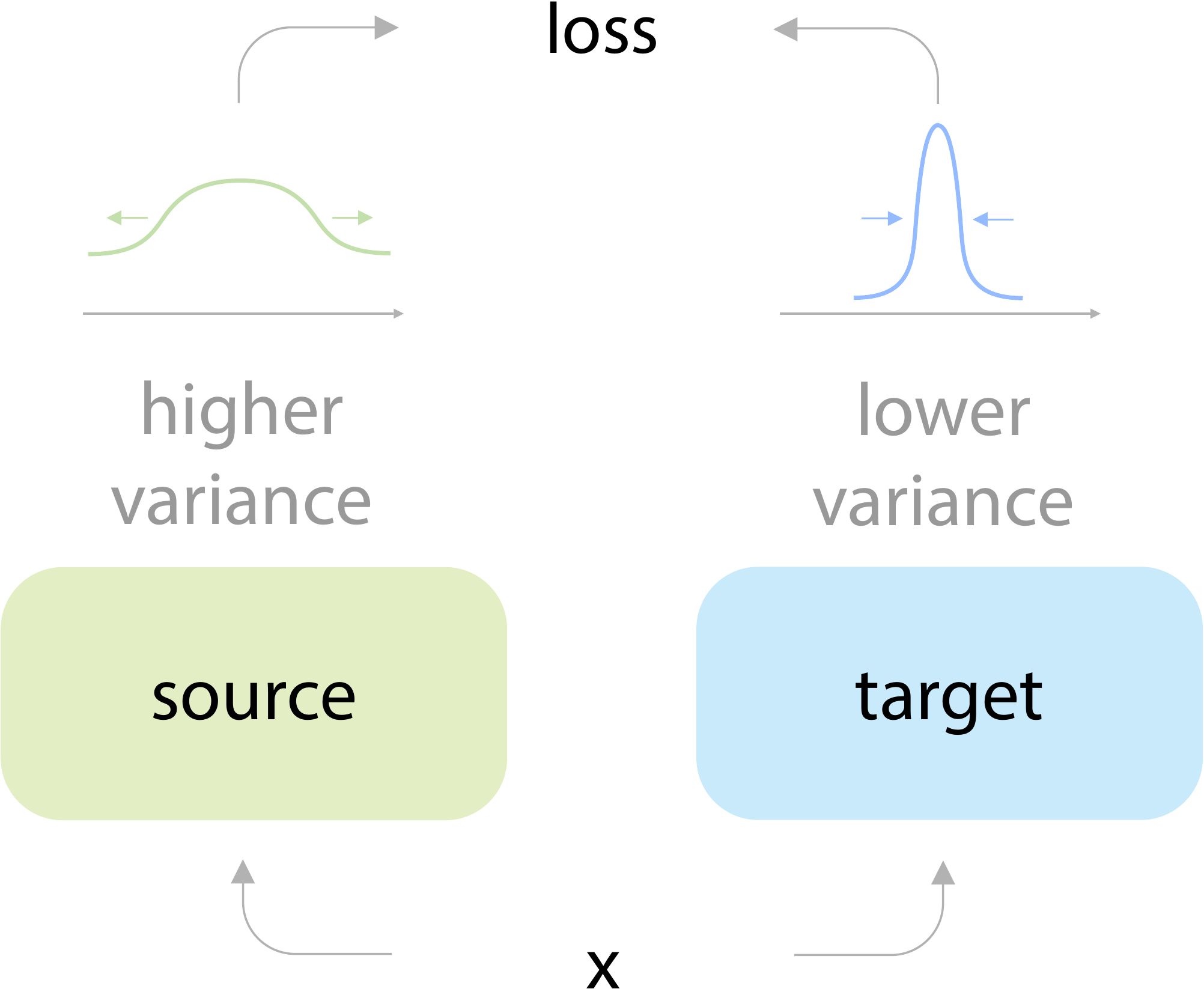}
  \caption{\textbf{Asymmetry for Siamese} representation learning. For the two encoders in a Siamese network, we treat one as a source encoder, and the other as a target encoder. We find it generally beneficial to have relatively lower variance in target than source.\label{fig:teaser}}
  \vspace{-.5em}
\end{figure}

In this paper, we conduct a more formal study on the importance of asymmetry for Siamese learning. Deviating from the original meaning of `Siamese', we explicitly mark the two encoders within the network functionally different: a \emph{source} encoder and a \emph{target} encoder.\footnote{Depending on the context, \emph{source} has also been referred as query/online/student; and \emph{target} as key/teacher in the literature~\cite{he2020momentum,grill2020bootstrap,tarvainen2017mean}.} The source encoder generates source encodings, and updates its weights via normal gradient-based optimization like in supervised learning. The target encoder updates its weights only with their source counterparts, and outputs target encodings which in turn judge the quality of sources. This asymmetric encoder formulation also covers symmetric encoders (\eg, in SimCLR~\cite{chen2020simple}), where the target weights can be simply viewed as source duplicates.

With this distinction, our key insight is that \emph{keeping a relatively lower variance in target encodings than source can help representation learning} (illustrated in \cref{fig:teaser}). We systematically study this phenomenon with our MoCo v2~\cite{chen2020improved} variant beyond existing -- but scattered -- evidence in the literature~\cite{he2020momentum,caron2020unsupervised,wang2021contrastive,cai2021exponential,li2021momentum}. Specifically, given a variance-oriented design, we first quantify its encoding variance with our baseline model, and then apply it to source or target (or both) encoders and examine the influence on learned representations. In total, we have conducted \emph{five} case studies to explore various design spaces, ranging from encoder inputs, to intermediate layers and all the way to network outputs. The results are well-aligned with our insight: designs that increase encoding variance generally help when applied to source encoders, whereas ones that decrease variance favor target. We additionally provide a preliminary theoretical analysis taking MoCo pre-training objective as an example, aimed at revealing the underlying cause. 

Our observation generalizes well. First, we show the improvements from asymmetry -- lower variance in target than source -- can hold with longer pre-training schedules, suggesting they are not simply an outcome of faster convergence. Second, directly applying proper asymmetric designs from MoCo v2 to a variety of other frameworks (\eg, BYOL~\cite{grill2020bootstrap}, Barlow Twins~\cite{zbontar2021barlow}) also works well, despite notable changes in objective function (contrastive or non-contrastive), model optimization (large-batch training~\cite{You2017} or not), \etc. Third, using MoCo v3~\cite{chen2021empirical}, we also experimented a more recent backbone -- Vision Transformer (ViT)~\cite{Dosovitskiy2021} -- and find the generalization still holds well. Finally, several asymmetric designs are fairly compositional: their combined effect enables single-node pre-trained MoCo v2 to reach a top-1 linear probing accuracy of \emph{75.6}\% on ImageNet, a state-of-the-art with ResNet-50 backbone. This model also demonstrates good transferring ability to other downstream classification tasks~\cite{chen2020simple,grill2020bootstrap,dwibedi2021little}.

In summary, our study reveals an intriguing correlation between the relative source-target variance and the learned representation quality. We have to note that such correlation has limitations, especially as self-supervised learning follows a staged evaluation paradigm and the final result is inevitably influenced by many \emph{other} factors. Nonetheless, we hope our exploration will raise the awareness of the important role played by asymmetry for Siamese representation learning, and inspire more research in this direction.

\section{Related Work\label{sec:related}}

\paragraph{Siamese networks} are weight-sharing networks~\cite{Bromley1994} that process multiple inputs and produce multiple outputs in parallel. It has been widely used in computer vision~\cite{Bromley1994,taigman2014deepface,wang2018iterative,bertinetto2016fully} and has recently caught attention in self-supervised learning~\cite{chen2020simple,chen2020exploring}. This can be explained by the design of Siamese networks, which can conveniently learn \emph{invariance} in a data-driven fashion -- a widely acknowledged property for useful visual representations~\cite{chen2020exploring}. While a na\"{i}ve application of Siamese network can incur collapse, various formulations and mechanisms (\eg, contrastive learning~\cite{he2020momentum,chen2020simple}, online balanced clustering~\cite{caron2020unsupervised,caron2021emerging}, extra predictor~\cite{grill2020bootstrap,chen2020exploring}, variance reduction loss~\cite{zbontar2021barlow,bardes2021vicreg}) -- many of them asymmetric -- have been proposed to maintain healthy learning dynamics. Our focus is \emph{not} on collapse prevention. Instead, we study generic designs that change encoding variance, analyze their effect on the output representations, and show that an asymmetry between source and target helps learning.

\paragraph{Symmetry for Siamese learning.} While the theme of the paper is asymmetry, symmetry is also a powerful concept in Siamese learning. One advantage of symmetry is in reducing the computation cost when source and target encoders share the same backbone weights. In such frameworks~\cite{chen2020simple,chen2020exploring}, source features can be \emph{reused} for targets, saving the extra need to compute with a second encoder. Recently, symmetric designs alone are also shown to yield the same level of performance as asymmetric methods~\cite{zbontar2021barlow,bardes2021vicreg}. 

Interestingly, there is often an attempt to \emph{symmetrize} the loss by forwarding image views once as source and once as target~\cite{grill2020bootstrap,chen2021empirical}, even when the encoder weights are \emph{not} shared (\eg, in case of a momentum encoder~\cite{he2020momentum}). Compared to using a single asymmetric loss but training for 2{\x} as long, this practice has the same number of forward/backward passes and we empirically verify it generates similar results across frameworks (see \cref{sec:frameworks})~\cite{chen2020exploring}. Therefore, we believe loss symmetrization is \emph{not} essential beyond plausible better performance at the `same' training epochs.

\paragraph{Asymmetric source-target variance.} Asymmetry in variance is already serving self-supervised learning in implicit ways. MoCo~\cite{he2020momentum} itself is a successful example: by smoothing its target encoder, the memory bank stores consistent keys with smaller variance across training iterations. Momentum update has been extended to normalization statistics to further reduce variance~\cite{cai2021exponential,li2021momentum}, again applied on targets. State-of-the-art on ImageNet~\cite{wang2021contrastive,zhou2021theory,xu2020seed} is held by using high-variance, strong augmentations on source views.

Siamese networks are also popular in \emph{semi}-supervised learning, where some examples are unlabeled. To create more reliable pseudo labels, the common practice is to average predicted labels over augmented views~\cite{berthelot2019mixmatch,wang2019enaet,sohn2020fixmatch}, which effectively reduces variance on target. 
Such evidences are scattered in the literature, and we analyze it systematically.

\begin{figure*}[t]
\centering
\subfloat[
\textbf{\mcrop} (\cref{sec:multicrop})
\label{fig:mcrop}
]{%
\centering
\begin{minipage}{0.195\linewidth}
\begin{center}
\includegraphics[width=0.85\linewidth]{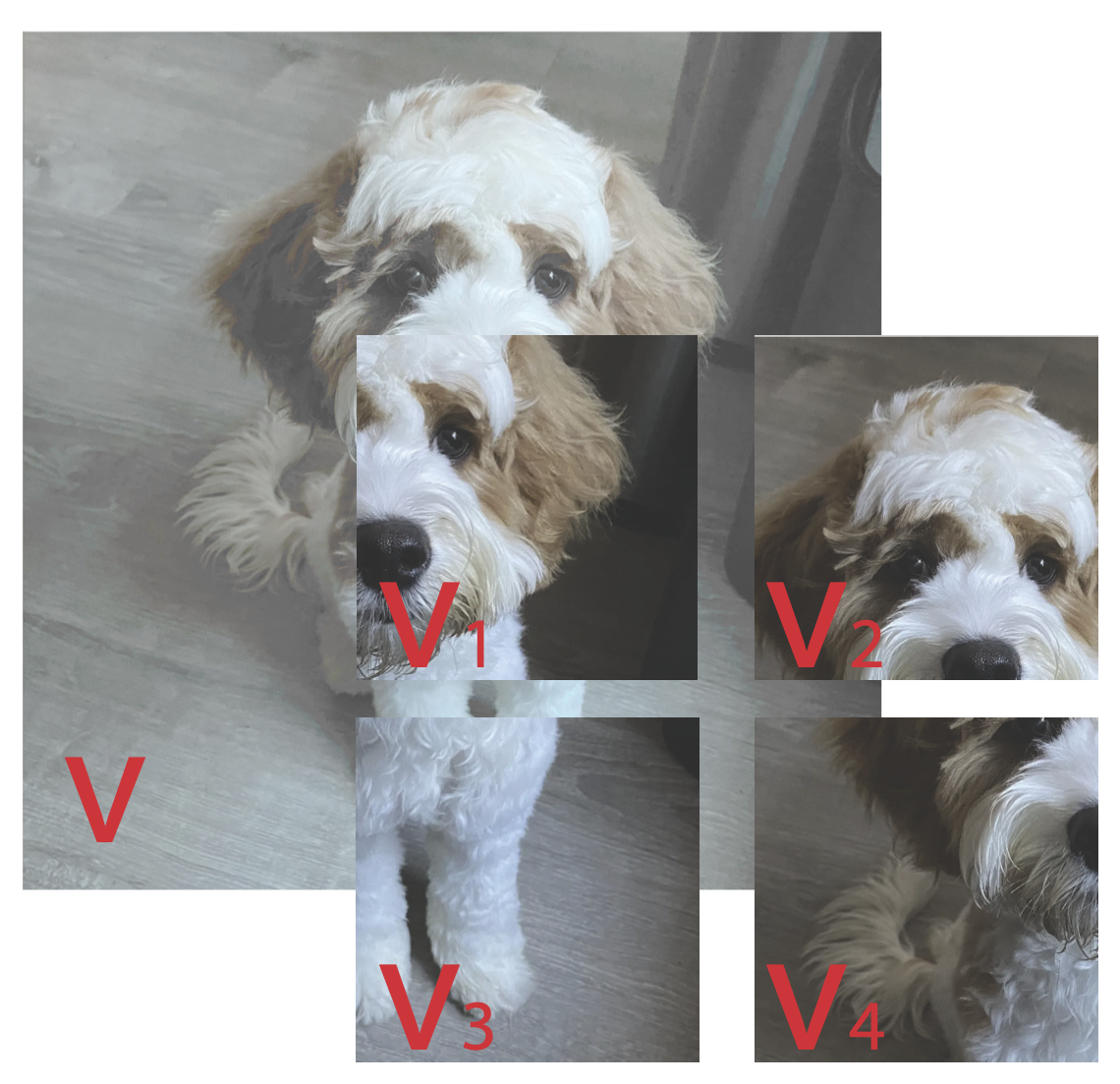}\\\vspace{.1em}
\includegraphics[width=1.0\linewidth]{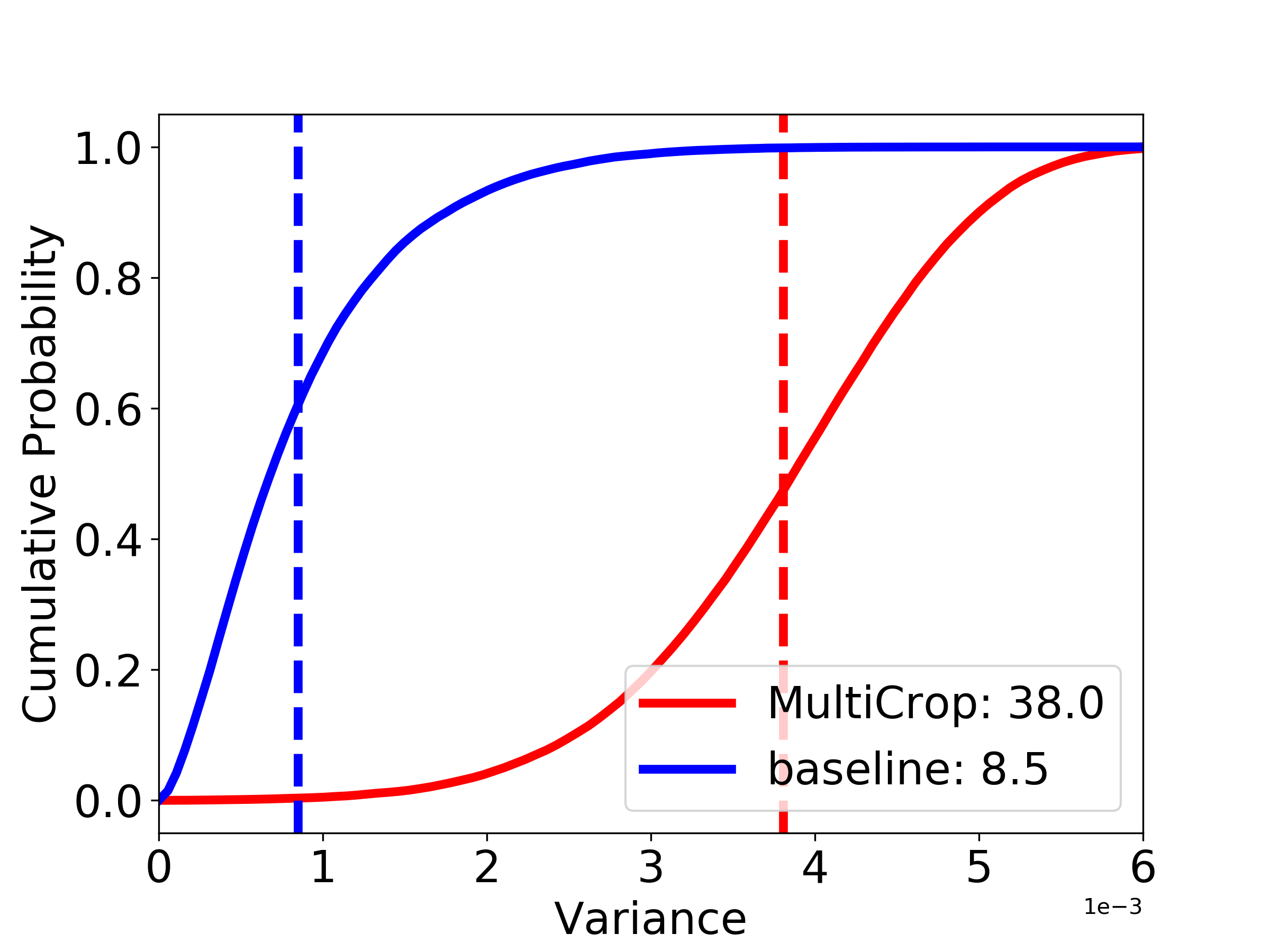}
\end{center}
\end{minipage}
}
\subfloat[
\textbf{\smix} (\cref{sec:scalemix})
\label{fig:scalemix}
]{%
\centering
\begin{minipage}{0.195\linewidth}
\begin{center}
\includegraphics[width=.85\linewidth]{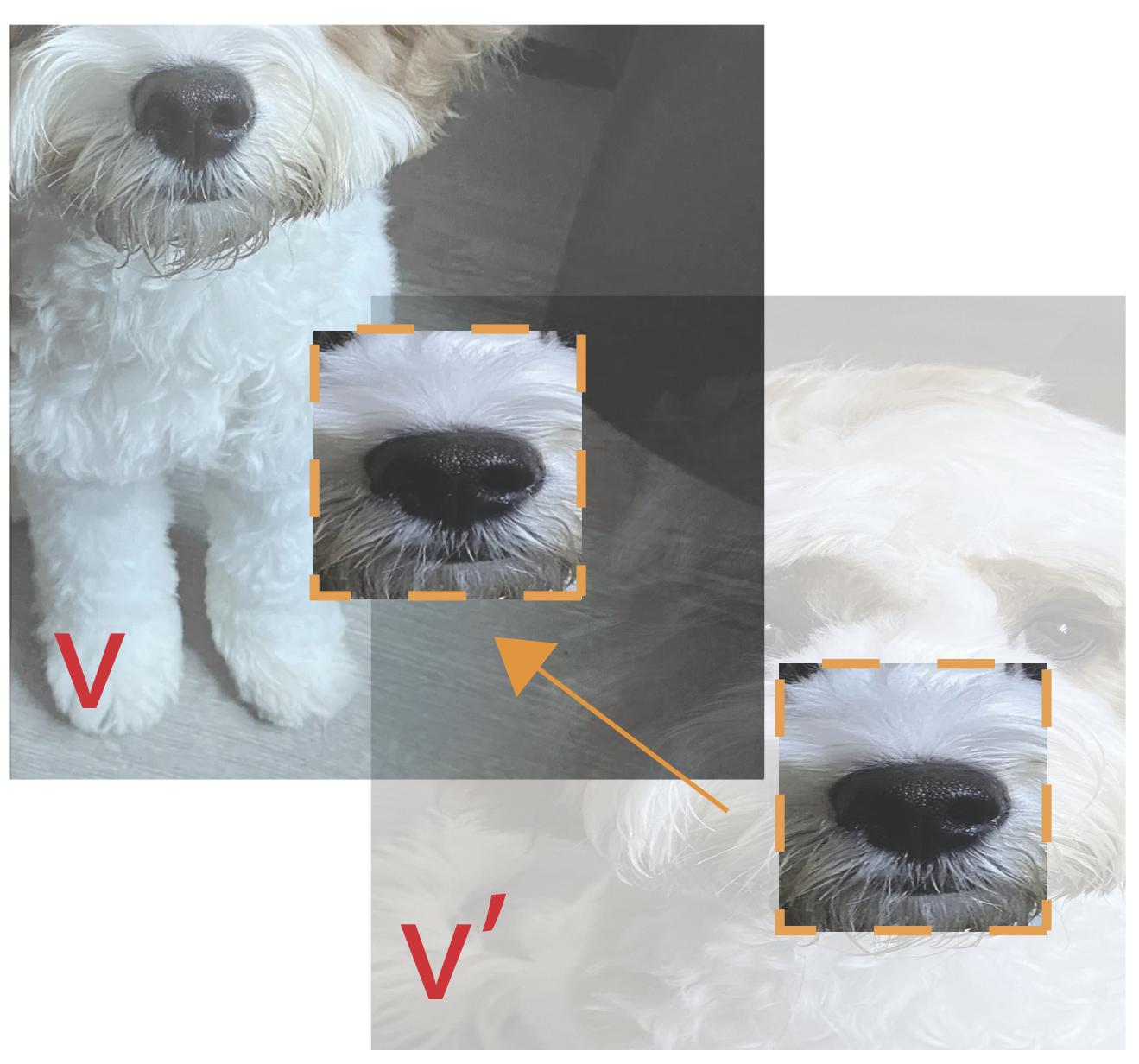}\\\vspace{.1em}
\includegraphics[width=1.0\linewidth]{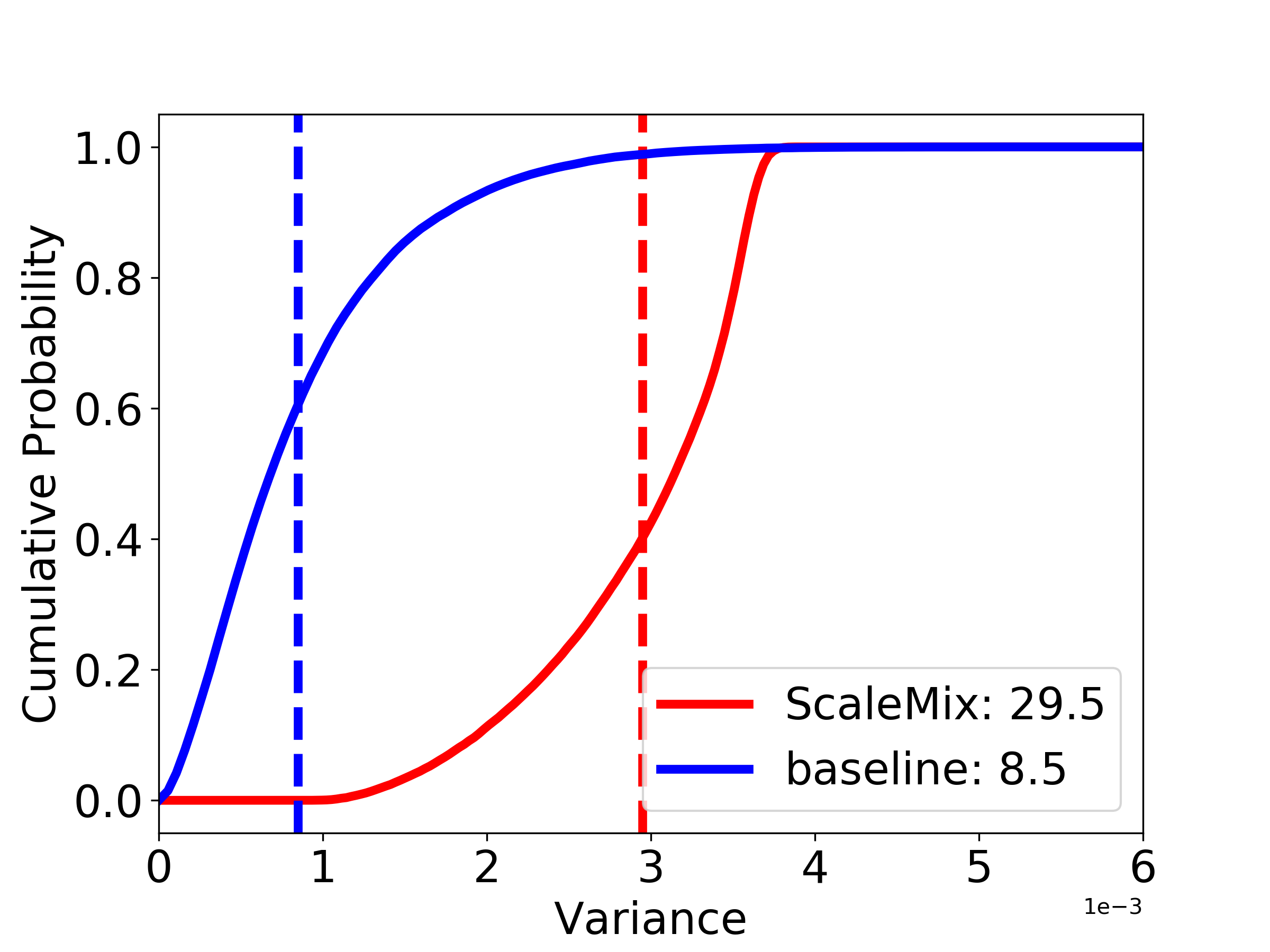}
\end{center}
\end{minipage}
}
\subfloat[
\textbf{\aaug} (\cref{sec:asymaug})
\label{fig:asym_aug}
]{%
\centering
\begin{minipage}{0.195\linewidth}
\begin{center}
\includegraphics[width=.85\linewidth]{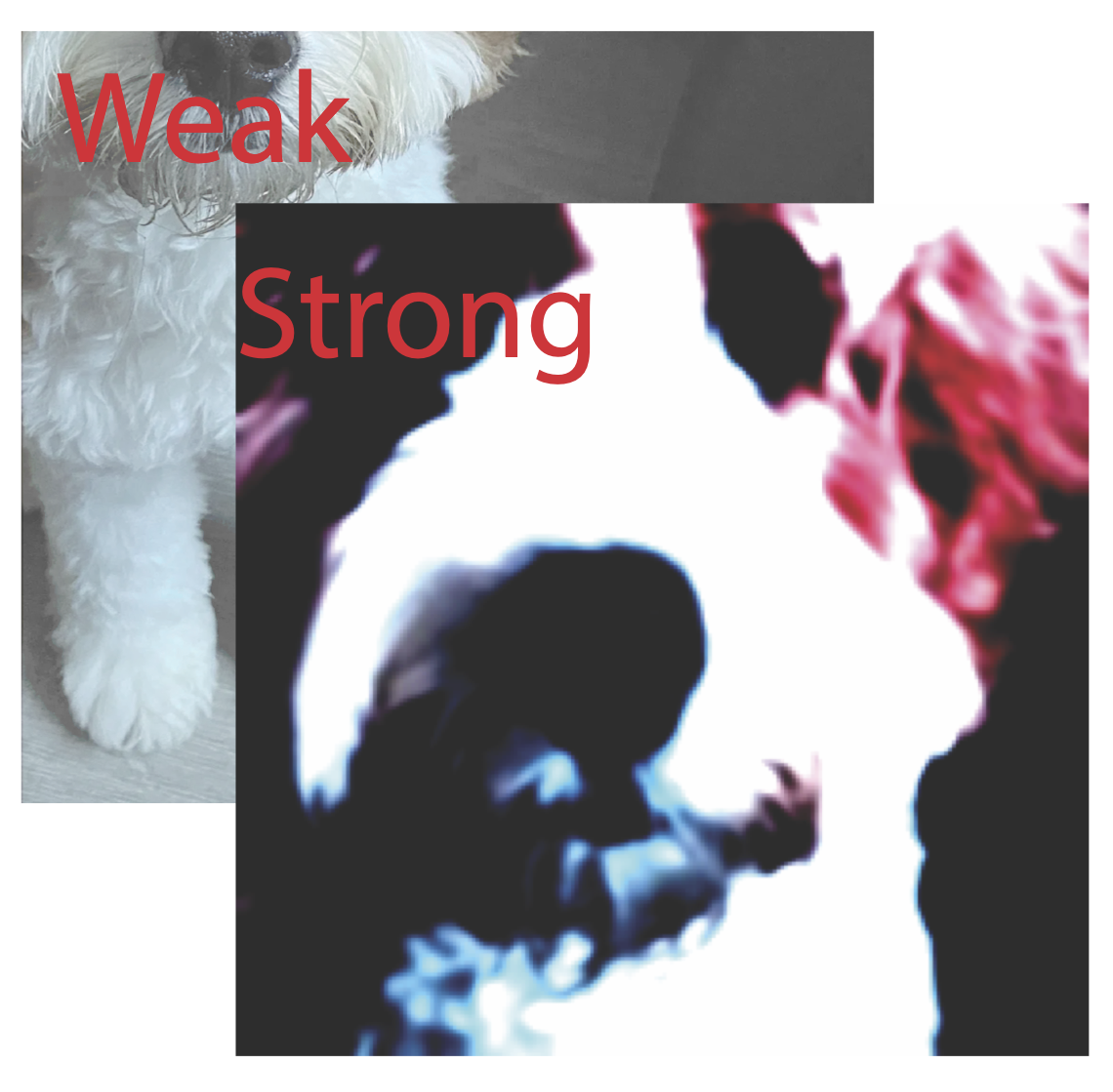}\\\vspace{.1em}
\includegraphics[width=1.0\linewidth]{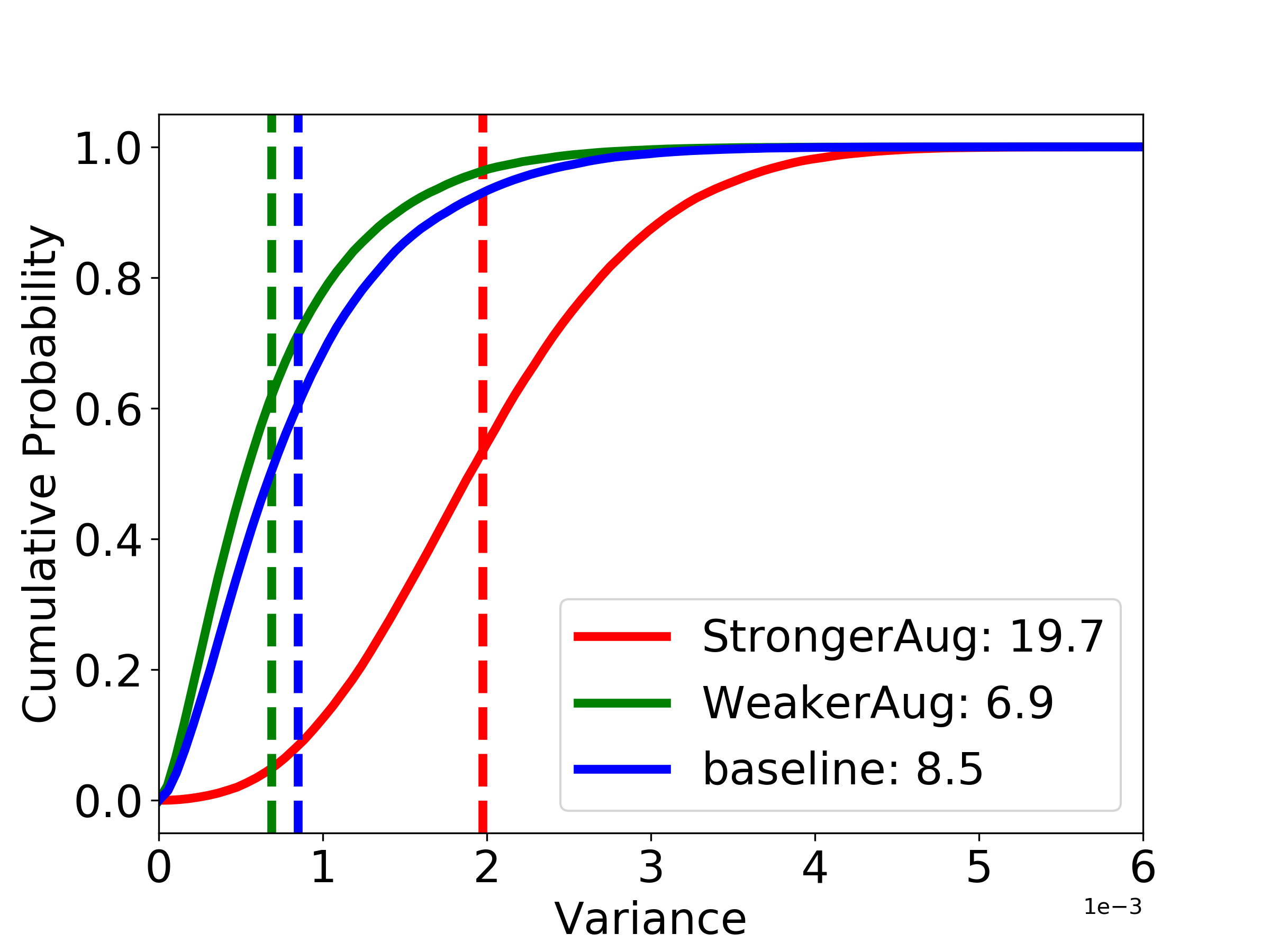}
\end{center}
\end{minipage}
}
\subfloat[
\textbf{\syncbn} (\cref{sec:syncbn})
\label{fig:asym_bn}
]{%
\centering
\begin{minipage}{0.195\linewidth}
\begin{center}
\hspace{-.1\linewidth}
\includegraphics[width=1.1\linewidth]{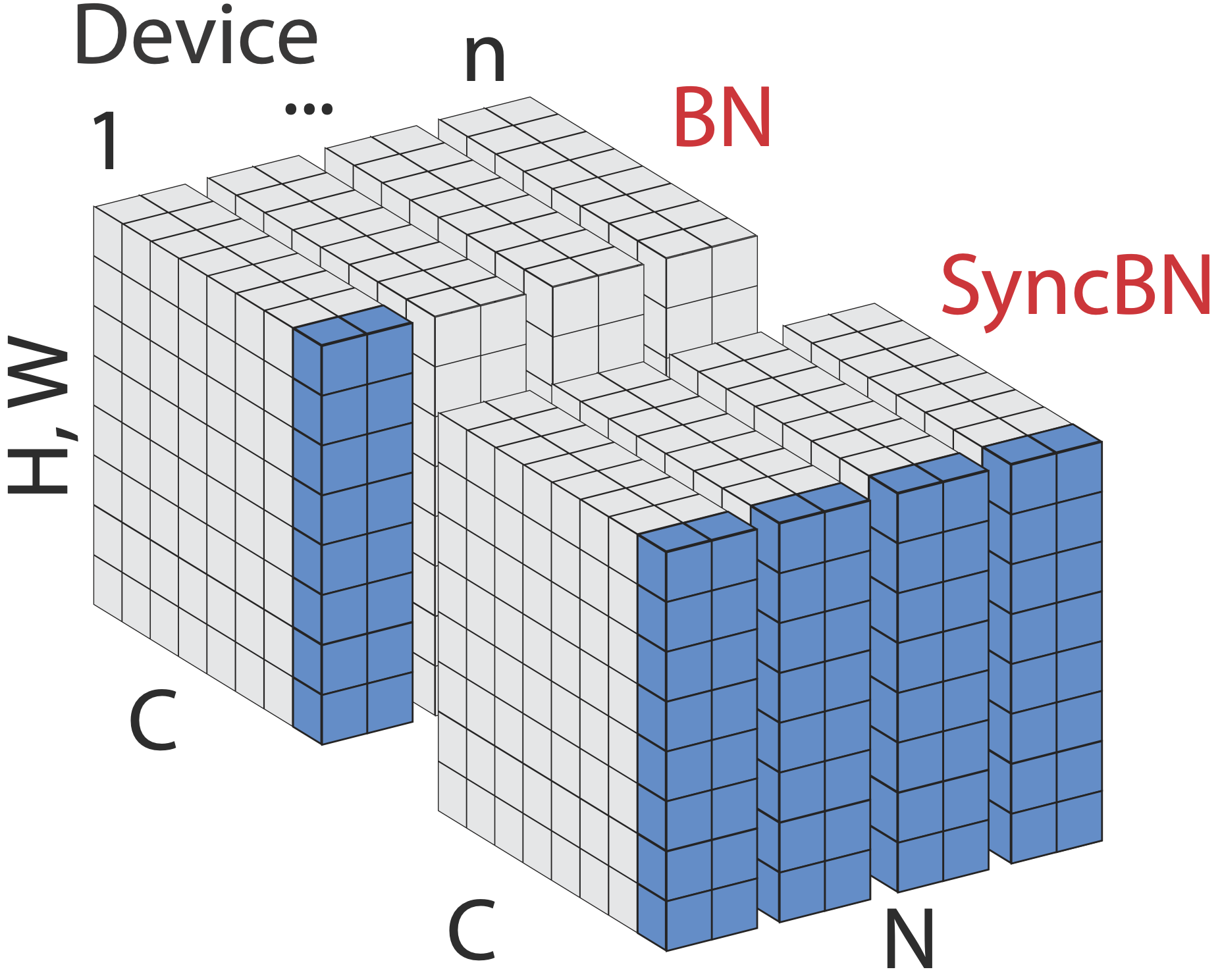}\\\vspace{.1em}
\includegraphics[width=1.0\linewidth]{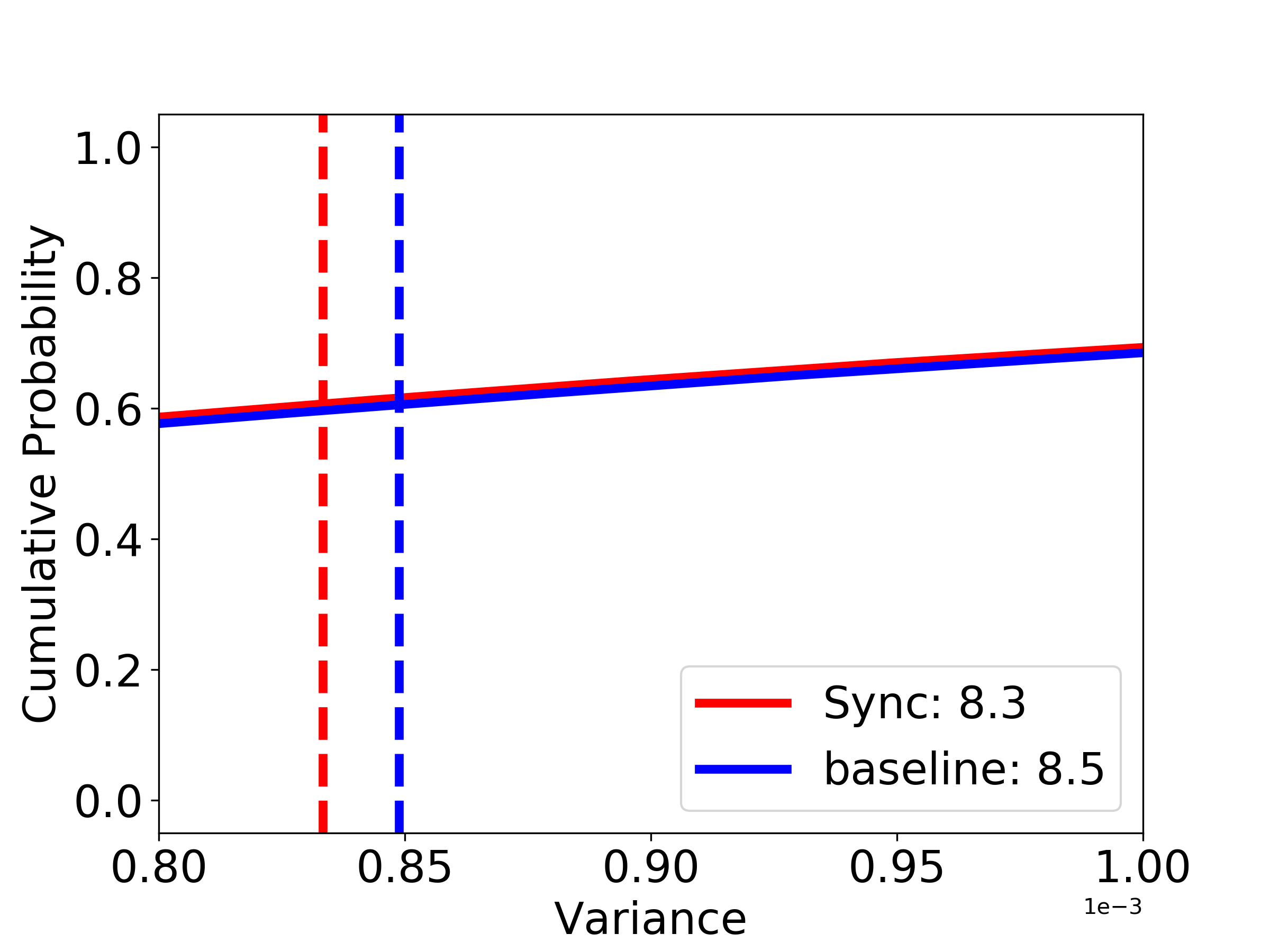}
\end{center}
\end{minipage}
}
\subfloat[
\textbf{\menc} (\cref{sec:mean_enc})
\label{fig:mean_enc}
]{%
\begin{minipage}{0.195\linewidth}
\begin{center}
\includegraphics[width=.8\linewidth]{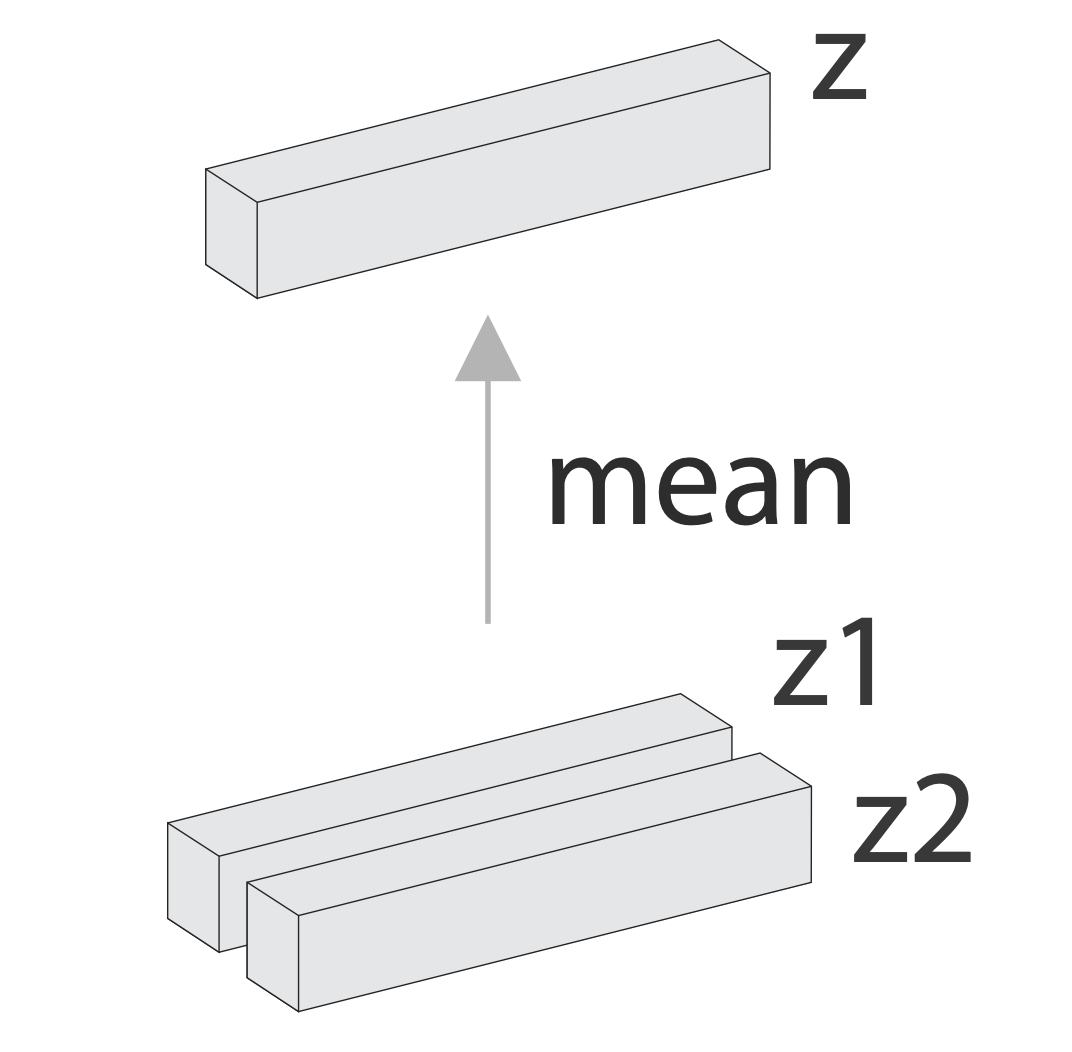}\\\vspace{0.1em}
\includegraphics[width=1.0\linewidth]{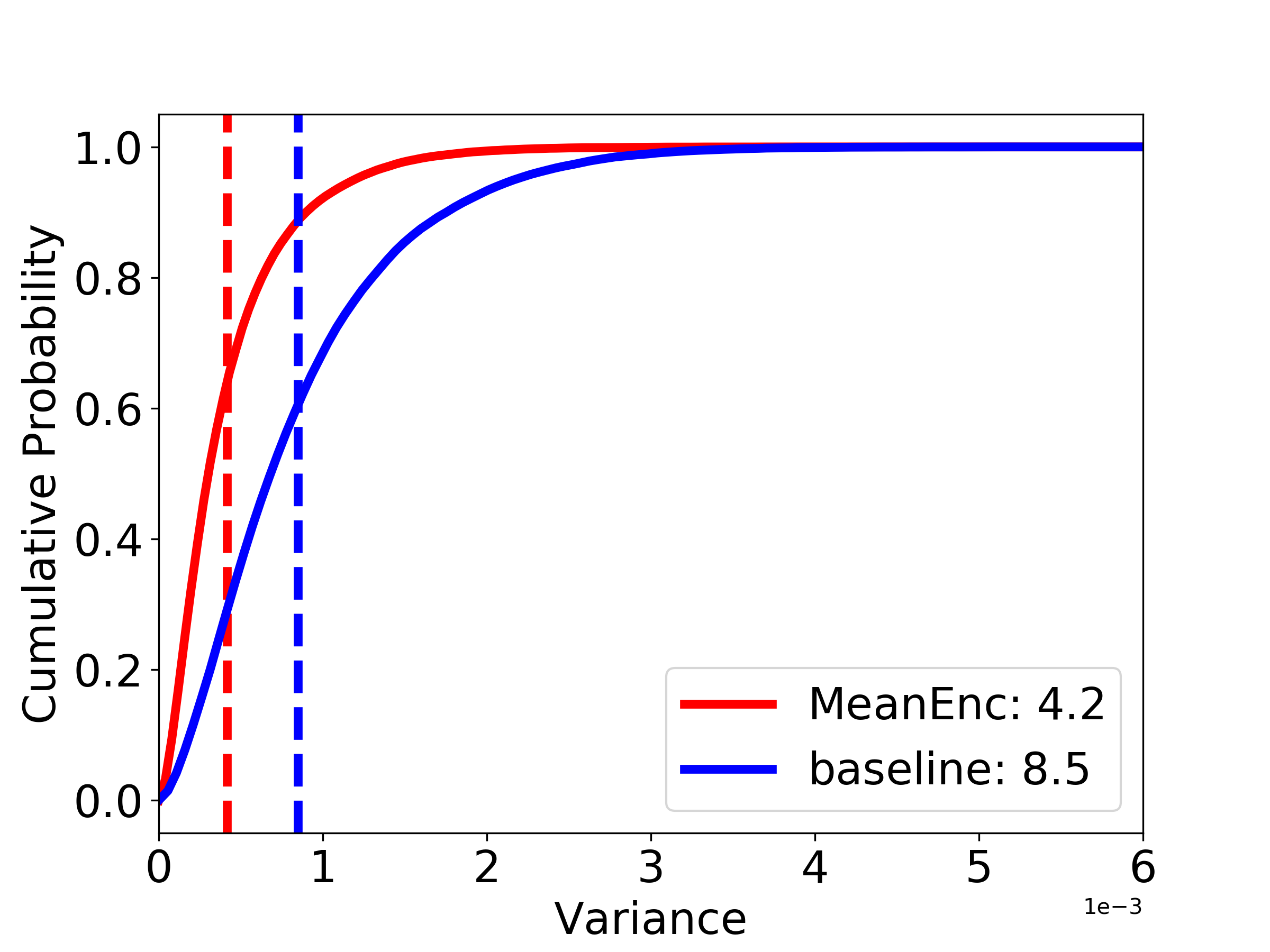}
\end{center}
\end{minipage}
}
\vspace{-.2em}
\caption{We present \textbf{five case studies} exploring different variance-oriented designs for source and target encoders. For each column, we show the specific design on the top, and its influence on the encoding variance (both the cumulative distribution function and the mean on the validation set as our empirical reference) at the bottom. Each design is then applied to either the source, the target, or both encoders. The resulting representation is evaluated by linear probing on ImageNet. Best viewed on a screen and zoomed in. See \cref{sec:study} for details.\label{fig:cases}}
\vspace{-1em}
\end{figure*}

\section{Methodology Overview\label{sec:overview}}

In this section we give an overview for our methodology to systematically study variance-oriented encoder designs. First, we specify our variance of interest. While exactly quantifying such variance during training is hard, we provide an approximate \emph{reference} for such variance using our baseline model. Now, for each design we can then compute its variance reference and quantify the \emph{relative} change in comparison to a vanilla encoder. Regardless of the change (higher or lower), we plug-in the design to either the source, the target, or both encoders and see its influence on resulting representations after pre-training. The influence is measured by linear probing on ImageNet~\cite{deng2009imagenet}. For a particular design, if applying it to both (or neither) encoders is better, then it implies maintaining symmetry is important; if it prefers either source or target, then it means \emph{asymmetry} is beneficial. In such cases, we also check whether the change in variance is correlated with the encoder preference.

In total, we have conducted \emph{five} case studies exploring various design spaces, ranging from encoder inputs (\ie, data augmentations), to intermediate layers (\ie, different batch sizes for Batch Normalization~\cite{ioffe2015batch}) all the way to network outputs (\ie, averaging multiple encodings to reduce variance). \cref{fig:cases} shows these designs and their variance plots in conjunction with our baseline. We detail our baseline and each case study in \cref{sec:study}, and first motivate our variance of interest and its reference in the following.

\paragraph{Variance of interest.} As each encoding is the encoder output of an augmented view from an image, the total variance in encodings mainly comes from three types: \emph{i)} changes to the encoder, \emph{ii)} changes across images, and \emph{iii)} changes within a single image. For type i), MoCo~\cite{he2020momentum} with its momentum encoder is already a major, well-studied asymmetric design that intuitively reduces the target variance across training iterations. For type ii), as Siamese representation learning encourages \emph{uniformity}~\cite{wang2020understanding,chen2020exploring}, the cross-image variance quickly converges to a constant dependent only on encoding dimensions (evidenced in \cref{sec:variance_appendix}).\footnote{If encodings are uniformly distributed on the unit hypersphere (due to $\ell_2$ normalization), their variance is $1/d$ where $d$ is the encoding dimension.} Therefore, we focus on type iii), \ie, \emph{intra}-image variance as the main subject of our study. Note that it does not restrict us to design input augmentations as the only means to adjust variance, as will be discussed in \cref{sec:syncbn,sec:mean_enc}.

\paragraph{Variance reference.} Exactly quantifying intra-image variance requires sampling all possible augmentations of all images and forward all of them to obtain encodings for all training steps. Even if possible, this process is highly expensive and also probably unnecessary. Therefore, we resort to an approximation with the goal of keeping a reference to characterize the encoding variance when changed.

To this end, we simply augment each image in the validation set $r$ times and feed them to a pre-trained baseline encoder. The output encodings are then used to compute the per-image, intra-sample variance, which jointly form a distribution. All variances across the entire set are then averaged to a single value $\varm$, the reference variance used to measure different designs. More details are listed in \cref{sec:details}.

\section{Case Studies for Source-Target Variance\label{sec:study}}

In this section, we introduce our baseline and perform five empirical case studies exploring the impact of different designs. For each one of them, we record its corresponding variance reference $\varm$, and linear-probing accuracies when placed on encoders with different configurations \emph{without preset bias}. Since our goal is to analyze the behavior, all models in this section are pre-trained for 100 epochs, with the generalization toward longer schedules deferred to \cref{sec:longer} after we draw the connection between variance change and encoder preference in \cref{sec:summarize}.

\paragraph{Baseline.} Our baseline is an improved variant of MoCo v2~\cite{chen2020improved}, which itself is an improved baseline over original MoCo~\cite{he2020momentum}. It consists of a gradient-updated source encoder \fsrc, a momentum-updated target encoder \ftgt, and an encoding-updated memory bank~\cite{wu2018unsupervised}. Inspired by SimCLR~\cite{chen2020simple}, each MoCo v2 encoder further uses a projection head (projector), which is a 2-layer MLP \emph{without} Batch Normalization (BN)~\cite{ioffe2015batch} in-between. Our baseline adds an additional fully connected layer (2048-\emph{d}, \emph{with} BN) before the 2-layer MLP. Inherited from MoCo v1, all BNs in \fsrc are performed per GPU device, and all BNs in \ftgt are shuffled~\cite{he2020momentum}. All the output encodings $\vecz$ are $\ell_2$ normalized to unit-length vectors before InfoNCE loss~\cite{oord2018representation}. We do \emph{not} employ any loss symmetrization~\cite{grill2020bootstrap,caron2020unsupervised} in this baseline, thus one source/target pair only contributes to the loss \emph{once}.

Compared to vanilla MoCo v2~\cite{chen2020improved}, our baseline is generally better in linear probing on ImageNet~\cite{deng2009imagenet} (detailed in \cref{sec:details}). The table below summarizes the top-1 accuracy (\%) using ResNet-50~\cite{he2016deep} and the same evaluation protocol:
\begin{center}
\vspace{-.2em}
\small
\tablestyle{2pt}{1.1}
\begin{tabular}{y{50}|x{40}x{40}x{40}x{40}}
 & 100 ep & 200 ep & 400 ep & 800 ep \\
\shline
MoCo v2~\cite{chen2020improved} & 64.7 & 67.9 & 69.6 & 70.7 \\
MoCo v2, \emph{ours} & 65.8 & 69.0 & 70.5 & 71.9 \\
\end{tabular}
\vspace{-.2em}
\end{center}
The improvement (\app 1 percent) is consistent across different number of training epochs. We also notice no degradation in object detection transfer on VOC~\cite{everingham2010pascal} -- \eg, achieving 57.4 mAP at 800 pre-training epochs, same as original~\cite{chen2020improved}.
The variance reference for our baseline $\varm_0$ is $8.5$ (\x$10^{-4}$).

\subsection{Study 1: \mcrop Augmentation\label{sec:multicrop}}

We begin our study with an existing design in the literature -- multi-crop augmentation (or `\mcrop')~\cite{misra2020self,caron2020unsupervised,caron2021emerging}. Besides the two basic views needed for Siamese learning, \mcrop takes additional views from each image per iteration. To alleviate the added computation cost, a common strategy is to have $m$ low-resolution crops (\eg, 96\x96~\cite{caron2020unsupervised}) instead of standard-resolution crops (224\x224) as added views (illustrated in \cref{fig:mcrop} top for $m{=}4$). As a side effect, inputting small crops can potentially increase the variance for an encoder due to the size and crop-distribution changes. This is confirmed in \cref{fig:mcrop} bottom, where we compare the variance distribution of \mcrop to our baseline on the ImageNet val set. We show the cumulative distribution function in solid lines with increasing per-image variances from left to right, and the mean variances $\varm$ and $\varm_0$ in dotted vertical lines. \mcrop has significantly higher variance than our baseline: $\varm{=}38.0$ \vs $8.5$ (\x$10^{-4}$).

We plug-in \mcrop to either the source, the target, or both encoders (detailed in \cref{sec:details_appendix}). The table below summarizes the corresponding top-1 accuracy and change ($\Delta$) to the baseline in linear probing:
\begin{center}
\vspace{-.2em}
\small
\tablestyle{2pt}{1.1}
\begin{tabular}{y{66}|x{36}a{36}x{36}x{36}}
+\mcrop (\colorbox{verylightgray}{\stronger}) & neither & \src & \tgt & both \\
\shline
accuracy (\%) & 65.8 & 69.9 & 57.1 & 61.7 \\
$\Delta$ (\%) & / & \better{+4.1} & \worse{-8.7} & \worse{-4.1} \\
\end{tabular}
\vspace{-.2em}
\end{center}
As a design that increases variance (indicated by `\colorbox{verylightgray}{\stronger}' in table), \mcrop improves the accuracy substantially (+4.1\%) when applied to the source encoder, and hurts when applied to the target. When applied to both, the performance also degenerates significantly (-4.1\%), even with more crops processed per training iteration than to source alone. These results indicate that the source encoder is the preferred place of applying \mcrop (column shaded in \colorbox{verylightgray}{gray}) -- which also matches the common protocols in the literature when multi-crop augmentation is used~\cite{misra2020self,caron2020unsupervised,caron2021emerging}.

\subsection{Study 2: ScaleMix Augmentation\label{sec:scalemix}}
Next, we introduce and study a different type of augmentation called `\smix', illustrated in \cref{fig:scalemix} top (more details are found in \cref{sec:scalemix_appendix}). As the name suggests, it generates new views of an image by mixing two views of potentially different scales together via binary masking. The masking strategy follows CutMix~\cite{yun2019}, where an entire region -- denoted by a box with randomly sampled coordinates -- is cropped and pasted. Unlike CutMix, \smix only operates on views from the \emph{same} image, and the output is a \emph{single} view of standard size (224\x224). This single view can be regarded as an efficient approximation of multiple crops in \mcrop, without the need to process small crops separately. Like \mcrop, \smix also introduces extra variance to the encoding space (as shown in \cref{fig:scalemix} bottom), with a mean variance of $\varm{=}29.5$ (\x$10^{-4}$).

Again, we apply \smix augmentation to the source, the target, or both encoders without preset preference. The results for linear probing are summarized in the table below:
\begin{center}
\vspace{-.2em}
\small
\tablestyle{2pt}{1.1}
\begin{tabular}{y{66}|x{36}a{36}x{36}x{36}}
+\smix (\colorbox{verylightgray}{\stronger}) & neither & \src & \tgt & both \\
\shline
accuracy (\%) & 65.8 & 67.3 & 52.8 & 64.8 \\
$\Delta$ (\%) & / & \better{+1.5} & \worse{-13.0} & \worse{-1.0} \\
\end{tabular}
\vspace{-.2em}
\end{center}
We observe a similar trend as the \mcrop case: \smix benefits source encoders, harms target encoders, and the effect neutralizes when applied to both. This suggests source encoder is again the preferred choice for \smix.

\subsection{Study 3: General Asymmetric Augmentations\label{sec:asymaug}}

\mcrop and \smix are mostly on geometric transformations of images. Next, we study the behavior by varying other ingredients in the MoCo v2 augmentation recipe. 

The original v2 recipe is \emph{symmetric}: the same set of augmentations (\eg, random resized cropping, color jittering~\cite{wu2018unsupervised}, blurring~\cite{chen2020simple}) is used for both source and target. In this case study, we add or remove augmentations (beyond geometric ones), and present two more recipes: one deemed stronger (`\saug'), and the other weaker (`\waug') compared to the original one (detailed in \cref{sec:details_appendix}). Together, they can form general \emph{asymmetric} augmentation recipes for source and target. Complying with the intuition, we find \saug has higher variance $19.7$ (\x$10^{-4}$), and \waug has lower variance $6.9$ (\x$10^{-4}$) \wrt to the baseline $\varm_0$ (shown in \cref{fig:asym_aug} bottom). 

The results are split into three tables for clarity. The influence of \waug is summarized first:
\begin{center}
\vspace{-.2em}
\small
\tablestyle{2pt}{1.1}
\begin{tabular}{y{66}|x{36}x{36}a{36}x{36}}
+\waug (\colorbox{verylightgray}{\weaker}) & neither & \src & \tgt & both \\
\shline
accuracy (\%) & 65.8 & 51.0 & 67.2 & 46.8 \\
$\Delta$ (\%) & / & \worse{-14.8} & \better{+1.4} & \worse{-19.0} \\
\end{tabular}
\vspace{-.2em}
\end{center}
Interestingly, the effect of \waug on source/target encoder is \emph{opposite} compared to the previous studies: it hurts source but helps target (referred as `\aaug'). A symmetric \waug on both does not work, suggesting the heavy reliance of Siamese learning on augmentation recipes~\cite{chen2020simple,grill2020bootstrap}. On the \saug side:
\begin{center}
\vspace{-.2em}
\small
\tablestyle{2pt}{1.1}
\begin{tabular}{y{66}|x{36}a{36}x{36}x{36}}
+\saug (\colorbox{verylightgray}{\stronger}) & neither & \src & \tgt & both \\
\shline
accuracy (\%) & 65.8 & 66.7 & 62.2 & 66.2 \\
$\Delta$ (\%) & / & \better{+0.9} & \worse{-3.6} & \better{+0.4} \\
\end{tabular}
\vspace{-.2em}
\end{center}
It helps most when used only on source, but harms accuracy when used only on target. For completeness, we also experimented changing augmentation strength in opposite directions for source and target:
\begin{center}
\vspace{-.2em}
\small
\tablestyle{2pt}{1}
\begin{tabular}{y{66}|x{76}x{76}}
Stronger \& Weaker & \src\colorbox{verylightgray}{\stronger} \tgt\colorbox{verylightgray}{\weaker} & \src\colorbox{verylightgray}{\weaker} \tgt\colorbox{verylightgray}{\stronger} \\
\shline
accuracy (\%) & 67.2 & 44.3 \\
$\Delta$ (\%) & \better{+1.4} & \worse{-21.5} \\
\end{tabular}
\vspace{-.2em}
\end{center}
Compared to having \waug on target alone (67.2\%), further adding \saug on source does not bring extra gains. In contrast, stronger augmentations on target and weaker augmentations on source results in the worst performance in all the cases we have studied.

\subsection{Study 4: Sync BatchNorm\label{sec:syncbn}}

Although input data augmentation is a major source of intra-image variance, it is \emph{not the only} cause of such variance within output encodings. One notable source lies in intermediate BN layers~\cite{ioffe2015batch}, a popular normalization technique in modern vision architectures~\cite{he2016deep}. During training, the statistics for BN are computed \emph{per-batch}, which means if other images within the batch are replaced, the output will likely change even if the current image stays the same. As a result, the magnitude of this variance is largely controlled by the \emph{batch size}: a sufficiently large size can provide nearly stable statistics, whereas for small batches (\eg, below 16) the estimation is generally less accurate~\cite{wu2018group}. For MoCo v2, its effective batch size is 32, because the default BN performs normalization only on the same device (256 images/8 GPUs).\footnote{MoCo v2 inherits MoCo v1 and uses `shuffled BN' in \ftgt. It shuffles the input to avoid cheating but the normalization still happens per-device.} A natural alternative is to employ \syncbn that normalizes over all devices, so the batch size is 256 (illustrated in \cref{fig:asym_bn} top for 4 devices). From the zoomed-in variance plot (\cref{fig:asym_bn} bottom), \syncbn leads to a slight decrease in variance from $8.5$ to $8.3$ (\x$10^{-4}$) in this case -- suggesting 32 is already sufficiently stable in our baseline.

For efficiency and generalizability, we replace the \emph{single} BN in our 3-layer projector with \syncbn.\footnote{Replacing \emph{all} BNs including ones in ResNet also exhibits the same pattern. Replacing BNs in projector only is noticeably faster, and generalizes to other BN-free backbones such as ViT~\cite{Dosovitskiy2021}.} As before, we tried different combinations on encoders and the results are:
\begin{center}
\vspace{-.2em}
\small
\tablestyle{2pt}{1.1}
\begin{tabular}{y{66}|x{36}x{36}a{36}x{36}}
+\syncbn (\colorbox{verylightgray}{\weaker}) & neither & \src & \tgt & both \\
\shline
accuracy (\%) & 65.8 & 64.7 & 66.5 & 66.0 \\
$\Delta$ (\%) & / & \worse{-0.9} & \better{+0.7} & \better{+0.2} \\
\end{tabular}
\vspace{-.2em}
\end{center}
Despite the seemly minor modification, \syncbn still leads to a notable improvement when applied to target (referred as `\abn') and degeneration to source. \syncbn on both encoders is at-par with the baseline per-device BNs.

\subsection{Study 5: Mean Encoding\label{sec:mean_enc}}

In this last study we focus on the encoder output. According to basic statistics, a direct approach to reduce the variance of a random variable is to perform \iid sampling multiple times and take the mean as the new variable. Specifically for $\varm$, we can reduce it by a factor of \app$n$ if the output encoding $\vecz$ is averaged from $n$ separate encodings $\{\vecz^1,\ldots,\vecz^n\}$ (illustrated in \cref{fig:mean_enc} top for $n{=}2$).\footnote{Here the reduction is approximate because we jointly forward multiple views which doubles or triples the batch size in BN; and encodings are further $\ell_2$ normalized before calculating $\varm$.} These encodings can be simply generated by running the same encoder on $n$ augmented views of the same image (detailed in \cref{sec:details_appendix}). For example, we show $\varm$ is $4.2$ (\x$10^{-4}$), about half of $\varm_0$ when two encodings are averaged in  \cref{fig:mean_enc} bottom. We name this design `\menc' for an encoder.

As discussed in our \cref{sec:related} (also shown in~\cite{chen2020exploring}), increasing the number of views per training iteration can lead to better performance \emph{by itself}. To minimize this effect, we conduct our main analysis of \menc by \emph{fixing} the total number of views to 4 per training iteration. The 4 views are split between source (\nsrc) and target (\ntgt) encoders, shown in the first 3 result columns below:
\begin{center}
\vspace{-.2em}
\small
\tablestyle{2pt}{1.1}
\begin{tabular}{y{66}|x{36}x{36}x{36}a{36}}
+\menc (\colorbox{verylightgray}{\weaker}) & \makecell{\nsrc${=}1$\\\ntgt${=}3$} & \makecell{\nsrc${=}2$\\\ntgt${=}2$} & \makecell{\nsrc${=}3$\\\ntgt${=}1$} & \makecell{\nsrc${=}1$\\\ntgt${=}2$} \\
\shline
accuracy (\%) & 67.9 & 67.1 & 59.9 & 67.5 \\
$\Delta$ (\%) & \better{+2.1} & \better{+1.3} & \worse{-5.9} & \better{+1.7} \\
\end{tabular}
\vspace{-.2em}
\end{center}

\begin{table*}[th]
\centering
\small
\tablestyle{5pt}{1.1}
\begin{tabular}{y{72}|x{45}x{45}x{45}x{45}x{45}x{45}}
& \makecell{\mcrop\\(\cref{sec:multicrop})} & \makecell{\smix\\(\cref{sec:scalemix})} & \makecell{\waug\\(\cref{sec:asymaug})} & \makecell{\saug\\(\cref{sec:asymaug})} & \makecell{\syncbn\\(\cref{sec:syncbn})} & \makecell{\menc\\(\cref{sec:mean_enc})} \\
\shline
variance change & \stronger & \stronger & \weaker & \stronger & \weaker & \weaker \\
encoder preference & \src & \src & \tgt & \src & \tgt & \tgt 
\end{tabular}
\vspace{-.5em}
\caption{\textbf{Summary} of the 6 designs covered in our case studies. For each design, we list its qualitative change in intra-image variance $\varm$, and its preferred encoder. We see a consistent pattern that higher-variance designs prefer source, whilst lower-variance ones prefer target.\label{tab:summarize}}
\vspace{-.7em}
\end{table*}
With more views in the target encoder (and simultaneously fewer views in source), we observe a trend for better accuracy. Having 2 views in both encoders still keeps symmetry, so its improvement over baseline (65.8\%) is an outcome of more views. For simplicity, we also experimented \menc with 2 views in the target encoder alone (last column). The result strikes a better balance between speed and accuracy, so we pick this setting as default for \menc.

\subsection{Summary of Studies\label{sec:summarize}}

In total, we covered 6 variance-oriented designs in the 5 case studies described above. Interestingly, none of them achieves best result when designs are symmetrically applied to both (or neither) encoders. Instead, all of them have a single \emph{preferred} encoder in the Siamese network. This phenomenon directly supports the importance of asymmetry for Siamese representation learning.

Moreover, we observe a consistent pattern: designs that introduce higher encoding variance generally help when placed on source encoders, whereas designs that decrease variance favor target encoders. We summarize the relation between: i) change of variance and ii) encoder preference in \cref{tab:summarize}. This is well-aligned with our insight: the specific asymmetry of a relatively lower variance in target encodings than source can benefit Siamese representation learning, and not the other way around.

From the results, we do have to note that such a pattern holds within a \emph{reasonable} range of $\varm$, and more extreme asymmetry does not always lead to better performance (\eg, when further increasing source augmentation strength while having \waug in target). Moreover, asymmetry is usually not the only factor in play for self-supervised frameworks; \emph{other factors} (\eg the number of views in \menc) can also influence the final outcome of our pipelines.

\section{Theoretical Analysis for Variance\label{sec:theory}}

Here we aim to provide a preliminary theoretical analysis for MoCo following~\cite{tian2020understanding,tian2021understanding} (More details in \cref{sec:theory_appendix}). Consider the following simplified InfoNCE objective:\footnote{We make two simplifications to InfoNCE~\cite{oord2018representation} by ignoring $\ell_2$ normalization and the positive term $\exp(\vecS_{ii'}/\tau)$ in the denominator~\cite{yeh2021decoupled}.}
\begin{equation}
    \mathcal{L} = -\frac{1}{N}\sum_{i=1}^N \log \frac{\exp(\vecS_{ii'}/\tau)}{\sum_{j\neq i} \exp(\vecS_{ij'}/\tau)}, \label{eq:loss}
\end{equation}
where $N$ is batch size, $\tau$ is temperature, $\vecS_{ii'}{=}\vecz^\top_i \vecz'_i$ and $\vecS_{ij'}{=}\vecz^\top_i\vecz'_j$ are pairwise similarities between source encodings $\vecz_i$ and targets $\vecz'_i$ (target weights and encodings all come with prime $'$). For MoCo, gradients are only back-propagated through the source $\vecz_i$, but \emph{not} $\vecz'_i$ or $\vecz'_j$.

Now, let's take the last linear layer immediately before $\vecz$ as an example for analysis. Let $\vecf$ be the input features of this layer, $W$ be its weight matrix (so $\vecz{=}W\vecf$), and denotes coefficients $\alpha_{ij'}{=}\exp(\vecS_{ij'}/\tau)/\sum_{k\neq i} \exp(\vecS_{ik'}/\tau)$, we can write the gradient flow of $W$ as:
\begin{equation}
    \frac{\dd \mathcal{L}}{\dd W} = W' \frac{1}{\tau N}\sum_{i=1}^N \sum_{j\ne i} \alpha_{ij'}(\vecf'_j - \vecf_i')\vecf_i^\top. 
    \label{eq:noise-gradient-short}
\end{equation}

To study the behavior of gradients especially \wrt our variance of interest, we can model intra-image variance as an \emph{additive noise} in $\vecf$ (and $\vecf'$) that affects training. Specifically, let $\tilde\vecf$ be the feature corresponding to the original image, we can assume:
\begin{itemize}
    \item Source features $\vecf_i {=} \tilde \vecf_i {+} \vece_{i}$, with $\ee[\vece_{i}]{=} \bar \vece$ and $\var[\vece_{i}] {=} \Sigma$; 
    \item Target side $\vecf'_i {=} \tilde \vecf_i {+} \vece'_{i}$, with $\ee[\vece'_{i}] {=}  \bar \vece'$ and $\var[\vece'_{i}] {=} \Sigma'$. 
\end{itemize}

$\ee[\cdot]$ computes expectation and $\var[\cdot]$ outputs variance. Note that $\tilde \vecf_i$ and $\tilde \vecf_j$ are from different images, while $\vece_{i}$, $\vece'_{i}$ and $\vece'_{j}$ model intra-sample variance that comes from multiple sources, \eg, input augmentations, BNs with different batch sizes (\cref{sec:syncbn}), \etc. Due to the independent augmentation process, these noises are modeled as independent of each other. 

Under such setting, we can arrive at the following result (detailed derivations in \cref{sec:theory_appendix}) to better understand our observation from a theoretical perspective:

\emph{Higher variance on the target side is not necessary and can be less stable.} With higher variance on the target side (\ie, $\Sigma'$ has larger eigenvalues), the variance of the gradient \wrt $W$, $\var[{\dd \mathcal{L}} {/} {\dd W}]$, will become larger without affecting its expectation $\ee[{\dd \mathcal{L}} {/} {\dd W}]$. Intuitively, this asymmetry comes from an asymmetric structure in~\cref{eq:noise-gradient-short}: there is a subtraction term ($\vecf'_j {-} \vecf_i'$) on the target side, but not on the source side ($\vecf_i$). To make the training dynamics more stable, maintaining a relative lower variance on the target side than source is preferred.

\begin{figure*}[t]
  \centering
  \includegraphics[width=.195\linewidth]{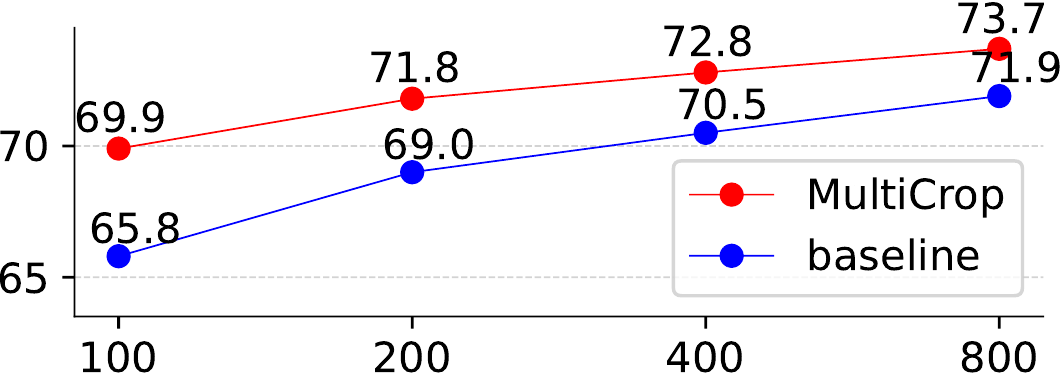}
  \includegraphics[width=.195\linewidth]{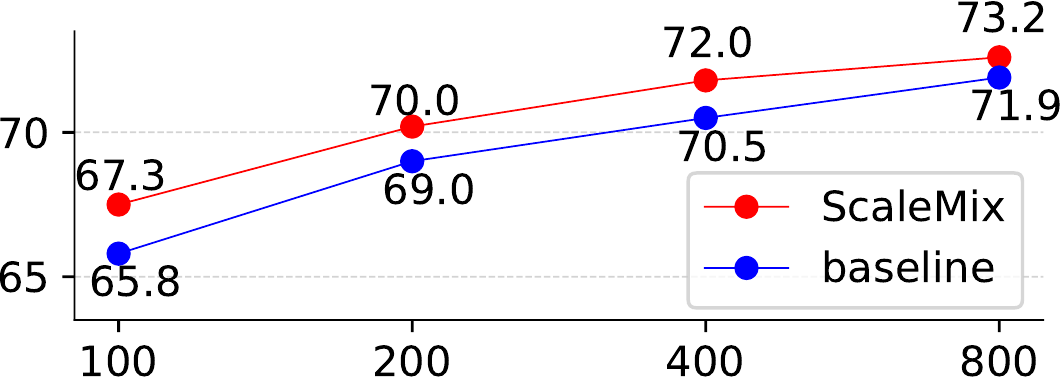}
  \includegraphics[width=.195\linewidth]{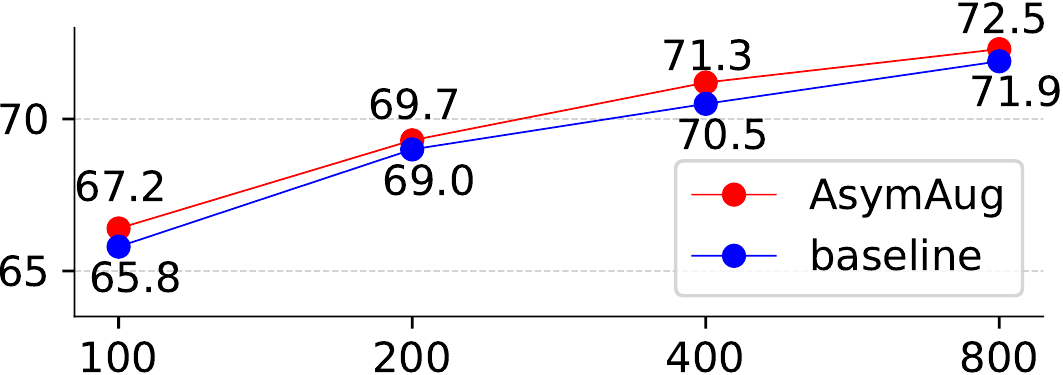}
  \includegraphics[width=.195\linewidth]{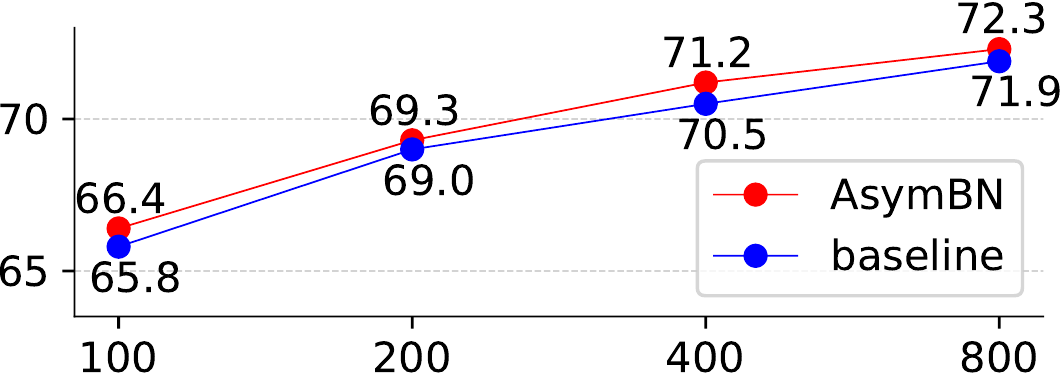}
  \includegraphics[width=.195\linewidth]{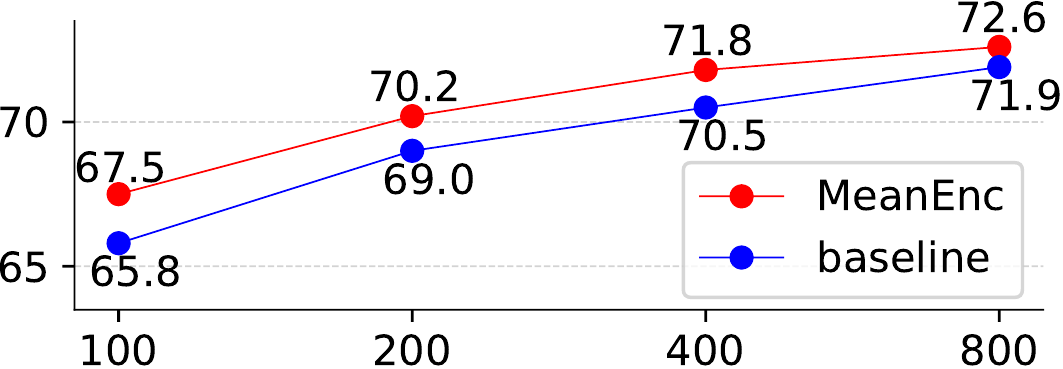}
  \vspace{-.2em}
  \caption{Generalization to \textbf{longer pre-training}. Here y-axis is accuracy (\%) and x-axis is number of epochs (log-scale). Asymmetric designs consistently outperform the baseline. \mcrop as the single strongest one reaches 73.7\% at 800-ep \emph{without} loss symmetrization. \label{fig:longer_train}}
  \vspace{-1em}
\end{figure*}

\section{Generalization Studies and Results\label{sec:generalize}}

The keyword of this section is \emph{generalization}, for which we study our insight for Siamese learning under various conditions. Specifically for MoCo v2, we study the behavior of asymmetric designs by training with longer schedules, and by composing multiple designs together. As a by-product, our final model achieves state-of-the-art on ImageNet, and performs well beyond when transferred to other datasets. Besides MoCo v2, we seek generalizations across more frameworks and backbones and find it also holds well. Unless otherwise specified, all the evaluations are top-1 linear probing accuracy on ImageNet~\cite{deng2009imagenet}.

\subsection{Longer Training\label{sec:longer}}

The first generalization is to longer training schedules. Most Siamese learning frameworks~\cite{chen2020simple,grill2020bootstrap,caron2020unsupervised}, including our baseline MoCo v2, produce substantially better results in linear probing with more training epochs. Meanwhile, lower variance in target -- in the extreme a \emph{fixed} target per image, could result in \emph{faster} convergence closer to supervised learning where longer training is not as helpful~\cite{he2016deep}. We run our baseline with the five asymmetric setups studied in \cref{sec:study} for 200, 400 and 800 epochs to check the behaviors, and put the trends in \cref{fig:longer_train}. Overall, \emph{all} the asymmetric models outperform the baseline across different epoch numbers. The maintained gap suggests the gain from asymmetry cannot be simply explained away by faster convergence.

\begin{table}[t]
\centering
\small
\tablestyle{2pt}{1.1}
\begin{tabular}{y{66}|x{36}|x{36}x{36}x{36}}
\makecell{(\%)} & baseline & \smix & \abn & \menc \\
\shline
MoCo v3~\cite{chen2021empirical} & \deemph{69.9} & \emph{70.7} & \emph{70.1} & \emph{70.6} \\
\emph{asym., 2\x} / $\Delta$ & \emph{69.7} & \better{+1.0} & \better{+0.4} & \better{+0.9} \\
\hline
SimCLR~\cite{chen2020simple} & \deemph{65.0} & \emph{66.3} & \emph{65.8} & \emph{66.4} \\
\emph{asym., 2\x} / $\Delta$ & \emph{65.0} & \better{+1.3} & \better{+0.8} & \better{+1.4} \\
\hline
BYOL~\cite{grill2020bootstrap} & \deemph{69.5} & \emph{70.4} & \emph{69.9} & \emph{69.7} \\
\emph{asym., 2\x} / $\Delta$ & \emph{69.0} & \better{+1.4} & \better{+0.9} & \better{+0.7} \\
\hline
SimSiam~\cite{chen2020exploring} & \deemph{67.8} & \emph{68.7} & \emph{68.0} & \emph{68.0} \\
\emph{asym., 2\x} / $\Delta$ & \emph{67.4} & \better{+1.3} & \better{+0.6} & \better{+0.6} \\
\hline
Barlow Twins~\cite{zbontar2021barlow} & \deemph{66.8} & \emph{67.3} & \emph{66.6} & \emph{67.1} \\
\emph{asym.} / $\Delta$ & \emph{66.4} & \better{+0.9} & \better{+0.2} & \better{+0.7} \\
\end{tabular}
\vspace{-.5em}
\caption{\label{tab:framework}Generalization to \textbf{more frameworks}. We cover \emph{5} of them and convert each to and \emph{asymmetric} one first. In the second column, we show similar results using our asymmetric versions compared to the original ones at 100-ep (in \textcolor{gray}{gray}), optionally with 2{\x} training schedules.\protect\footnotemark{} On top of these, we find asymmetric designs help learning across the board: third to fifth columns list accuracies and improvements over the asymmetric baseline.}
\addtocounter{footnote}{-1}
\vspace{-1em}
\end{table}

\subsection{More Frameworks\label{sec:frameworks}}

Next we examine the generalization to other frameworks. Roughly ranked by its similarity to our baseline MoCo v2 from closest to furthest, they are: i) \emph{MoCo v3}~\cite{chen2021empirical}, where the memory bank is replaced by large batch sizes~\cite{You2017}; ii) \emph{SimCLR}~\cite{chen2020simple}, where no momentum encoder is needed; iii) \emph{BYOL}~\cite{grill2020bootstrap}, where the contrastive formulation is challenged by learning only on comparing positive pairs; iv) \emph{SimSiam}~\cite{chen2020exploring}, where neither momentum encoder nor negative pairs are required; and v) \emph{Barlow Twins}~\cite{zbontar2021barlow}, where a fully symmetric pipeline for Siamese learning is discovered. Note that we only outlined major differences above and more subtleties (including detailed setup for each framework in this paper) are found in \cref{sec:details_appendix}.

For ease of applying asymmetric designs to these frameworks, we first convert their \emph{symmetrized} components to an asymmetric form following our source-target formulation. A popular one is loss symmetrization, used by all except Barlow Twins. We remove it by only forwarding a pair of views through the network \emph{once} (instead of \emph{twice}) per iteration. Intuitively, training 2{\x} as long can roughly compensate for the symmetrized loss with fair amount of compute, as discussed in \cref{sec:related} and analyzed in~\cite{chen2020exploring}. Moreover, methods without momentum encoders~\cite{chen2020simple,chen2020exploring,zbontar2021barlow} reuse source encoders for targets. In such cases, we explicitly maintain a target encoder by using an online clone of the source one, and \emph{stopping gradients} from flowing into the branch -- a choice deviated from SimCLR and Barlow Twins~\cite{chen2020simple,zbontar2021barlow}. We show in \cref{tab:framework} (second column) that our asymmetric versions work similarly in accuracy compared to the original ones, despite the above modifications.\footnote{We keep all the optimization hyper-parameters the same when running the asymmetric version. The results can be further improved when \eg learning rate is adjusted following the batch size change~\cite{Goyal2017}.}

We pick \smix, \abn and \menc as three representative designs which range from encoder inputs to outputs. \mcrop is relatively well studied in the literature~\cite{caron2020unsupervised,caron2021emerging} and we find it non-trivial to train \mcrop with large batch sizes~\cite{chen2021empirical,chen2020simple,grill2020bootstrap,zbontar2021barlow}. More recent frameworks~\cite{chen2021empirical,grill2020bootstrap,zbontar2021barlow} already employ stronger \emph{asymmetric} augmentation recipes~\cite{grill2020bootstrap} like \aaug. Thus we did not include them in our comparisons listed in \cref{tab:framework} (last three columns). Our asymmetric source-target designs generalize well beyond MoCo v2, showing consistent improvements across the board with same number of pre-training epochs.

\begin{table}[t]
\centering
\small
\tablestyle{2pt}{1.1}
\begin{tabular}{y{66}|x{36}|x{36}x{36}x{36}}
\makecell{(\%)} & baseline & \smix & \abn & \menc \\
\shline
MoCo v3, ViT~\cite{chen2021empirical} & \deemph{69.1} & \emph{69.1} & \emph{69.4} & \emph{69.4} \\
\emph{asym., 2\x} / $\Delta$ & \emph{68.7} & \better{+0.4} & \better{+0.7} & \better{+0.7} \\
\end{tabular}
\vspace{-.5em}
\caption{Generalization to \textbf{ViT}~\cite{Dosovitskiy2021}, a new architecture gaining popularity in vision and is recently studied in MoCo v3~\cite{chen2021empirical}. The procedure and table format follow \cref{tab:framework}. \label{tab:architecture}}
\vspace{-1em}
\end{table}

\begin{table*}[t]
\centering
\small
\tablestyle{2pt}{1.1}
\begin{tabular}{y{55}|x{32}x{34}x{38}x{32}x{32}x{32}x{26}x{32}x{26}x{26}x{40}x{32}}
                    & Food-101 & CIFAR-10 & CIFAR-100 & Birdsnap & SUN-397 & Cars & Aircraft & VOC-07 & DTD & Pets & Caltech-101 & Flowers \\                 
\shline
Supervised & 72.3 & 93.6 & 78.3 & 53.7 & 61.9 & 66.7 & 61.0 & \underline{87.5} & 74.9 & 91.5 & \textbf{94.5} & 94.7 \\
SimCLR~\cite{chen2020simple} &  68.4 & 90.6 & 71.6 & 37.4 & 58.8 & 50.3 & 50.3 & \underline{85.5} & 74.5 & 83.6 & 90.3 & 91.2 \\
BYOL~\cite{grill2020bootstrap} & 75.3 & 91.3 & 78.4 & 57.2 & 62.2 & 67.8 & 60.6 & 82.5 & 75.5 & 90.4 & 94.2 & \textbf{96.1} \\
NNCLR~\cite{dwibedi2021little} & 76.7 & \textbf{93.7} & \textbf{79.0} & \textbf{61.4} & 62.5 & 67.1 & \textbf{64.1} & 83.0 & 75.5 & \textbf{91.8} & 91.3 & 95.1 \\
\hline
Ours, 1600-ep & \textbf{79.4} & 92.8 & 77.8 & 58.5 & \textbf{67.8} & \textbf{69.7} & 59.3 & \underline{\textbf{93.8}} & \textbf{80.2} & 87.2 & 93.1 & 92.5 \\
\end{tabular}   
\vspace{-.7em}
\caption{Generalization by \textbf{transferring our model} to 12 different downstream datasets with linear probing. We follow the protocol of~\cite{grill2020bootstrap,dwibedi2021little} and report results on the test set. For VOC-07, we cite the improved numbers from~\cite{zbontar2021barlow} for fair comparisons. Our 1600-ep model achieves best results on 5 out of 12, while being less competitive on tasks with iconic images (such as CIFAR~\cite{Krizhevsky2009} and Aircraft~\cite{maji2013fine}).\label{tab:transfer}}
\vspace{-1.5em}
\end{table*}

\subsection{ViT Backbone\label{sec:vit}}
With MoCo v3, we also benchmarked a newly proposed backbone: ViT~\cite{Dosovitskiy2021}. We follow the same procedure by first building an asymmetric baseline and then applying different designs (detailed in \cref{sec:details_appendix}). Again, we find asymmetry works well (\cref{tab:architecture}). The only notable difference is the reduced gap for \smix, which is likely related to patches fed for ViT \emph{not aligned} with ScaleMix masks~\cite{jiang2021all}.

\subsection{Design Compositions\label{sec:final_res}}
As another aspect for generalization, we compose multiple asymmetric designs together and check their joint effect on representation quality. To this end, we fall back to our MoCo v2 baseline (100-ep) and start from our strongest single asymmetric design, \mcrop. When pairing it with other two input designs (\smix an \aaug), we find their added value has mostly diminished so we did not include them. On the target side, we first enabled \syncbn, and then enabled \menc (\ntgt${=}2$) to reduce variance, and both designs further improved performance:
\begin{center}
\vspace{-.2em}
\small
\tablestyle{2pt}{1.1}
\begin{tabular}{y{50}|x{36}x{38}x{42}a{44}}
\vspace{.35em}compositions & none & +\mcrop & +\mcrop+\abn & +\mcrop+\abn+\menc \\
\shline
accuracy (\%) & 65.8 & 69.9 & 70.4 & 71.3 \\
$\Delta$ (\%) & -&\better{+4.1} & \better{+4.6} &\better{+5.5} \\
\end{tabular}
\vspace{-.2em}
\end{center}
While our exploration on this front is preliminary and improvement is not guaranteed (as discussed in~\cref{sec:summarize}), it indicates different asymmetric designs can be compositional.

Finally, we pre-train our best composition (shaded column above) for 1600 epochs to check its limit. We arrive at \emph{75.6}\% on ImageNet linear probing (more details in \cref{sec:details}). This puts us in the state-of-the-art cohort~\cite{wang2021contrastive,zhou2021theory,xu2020seed} with single-node training and no other bells or whistles.

\subsection{Transfer Learning\label{sec:transfer}}
In \cref{tab:transfer}, we show transfer learning results of our final ImageNet 1600-ep model to 12 standard downstream classification tasks for linear probing~\cite{chen2020simple,grill2020bootstrap,dwibedi2021little}. For each dataset, we search the learning rate on the validation set and report results on the test set, following the protocol of~\cite{grill2020bootstrap,dwibedi2021little} (see \cref{sec:details_appendix}). Our model performs competitively against the most recent NNCLR~\cite{dwibedi2021little}), achieving best on 5 tasks but lags behind on ones with iconic images. We hypothesis it's due to \mcrop which used local small crops. We further transferred to Places-205~\cite{zhou2014learning}, which focuses on scene-level understanding. We find our model indeed achieves state-of-the-art (56.8\%), slightly better than SwAV~\cite{caron2020unsupervised} which also used \mcrop. These results verify our learned representation is effective beyond ImageNet.

\section{Implementation Details\label{sec:details}}
We list the most important implementation details for our paper below. Other subtleties are found in \cref{sec:details_appendix}.

\paragraph{Variance reference.} We use ImageNet val set (50k images in total), $r{=}32$ views, and the 800-ep pre-trained baseline source encoder for variance calculation.\footnote{A potential concern is the variance reference being biased by out-of-distribution views, since the baseline model has not seen certain data (\eg., small crops) during training. To address this, we also experimented with a model pre-trained with all the asymmetric designs. The trends still hold.} Encodings are $\ell_2$ normalized. To fully mimic the pre-training setting, we use online \emph{per-batch} statistics for BN, not recorded moving-average ones from the training set.

\paragraph{Pre-training.} By default, we adopt the same MoCo v2 setup (\eg, augmentation recipe, SGD optimizer \etc) for experiments on our baseline. A half-cycle cosine learning rate decay schedule~\cite{loshchilov2016sgdr} is used given the number of pre-training epochs. Mixed-precision is enabled for efficiency.

\paragraph{Linear probing.} Linear probing freezes backbone after pre-training, and only trains a linear classifier on top of the global image features to test the representation quality. By default on ImageNet, we use LARS~\cite{You2017} optimizer with batch size 4096, initial learning rate $lr{=}$1.6 (linearly scaled~\cite{Goyal2017}), weight decay 0 and train the classifier for 90 epochs with a half-cycle cosine schedule following SimSiam~\cite{chen2020exploring}. We choose LARS over SGD as the former shows better adaptation for explorations, without the need to search hyper-parameters (\eg $lr$) extensively for good performance. For our final model, we switched back to SGD optimizer following MoCo~\cite{he2016deep}, with an initial learning rate of 120 and batch size of 256.

\section{Conclusion\label{sec:conclusion}}
Through systematic studies, we have revealed an interesting correlation between the \emph{asymmetry} of source-target variance and the representation quality for Siamese learning methods. While such a correlation is conditioned on other factors and certainly not universal, we find as guideline it is generally applicable to various training schedules, frameworks and backbones. Composing asymmetric designs helps us achieve state-of-the-art with MoCo v2, and the learned representation transfers well to other downstream classification tasks. We hope our work will inspire more research exploiting the importance of asymmetry for Siamese learning, \eg for object detection transfer~\cite{he2020momentum} or speeding up model convergence for carbon neutral training.

\clearpage
\paragraph{Acknowledgements.} XC would like to thank Kaiming He on helpful discussions through this project. XW would like to thank Yutong Bai on helpful discussions through this project.

\appendix

\section{Cross-Image Variance\label{sec:variance_appendix}}
\begin{figure}[t]
  \centering
  \includegraphics[width=.95\linewidth]{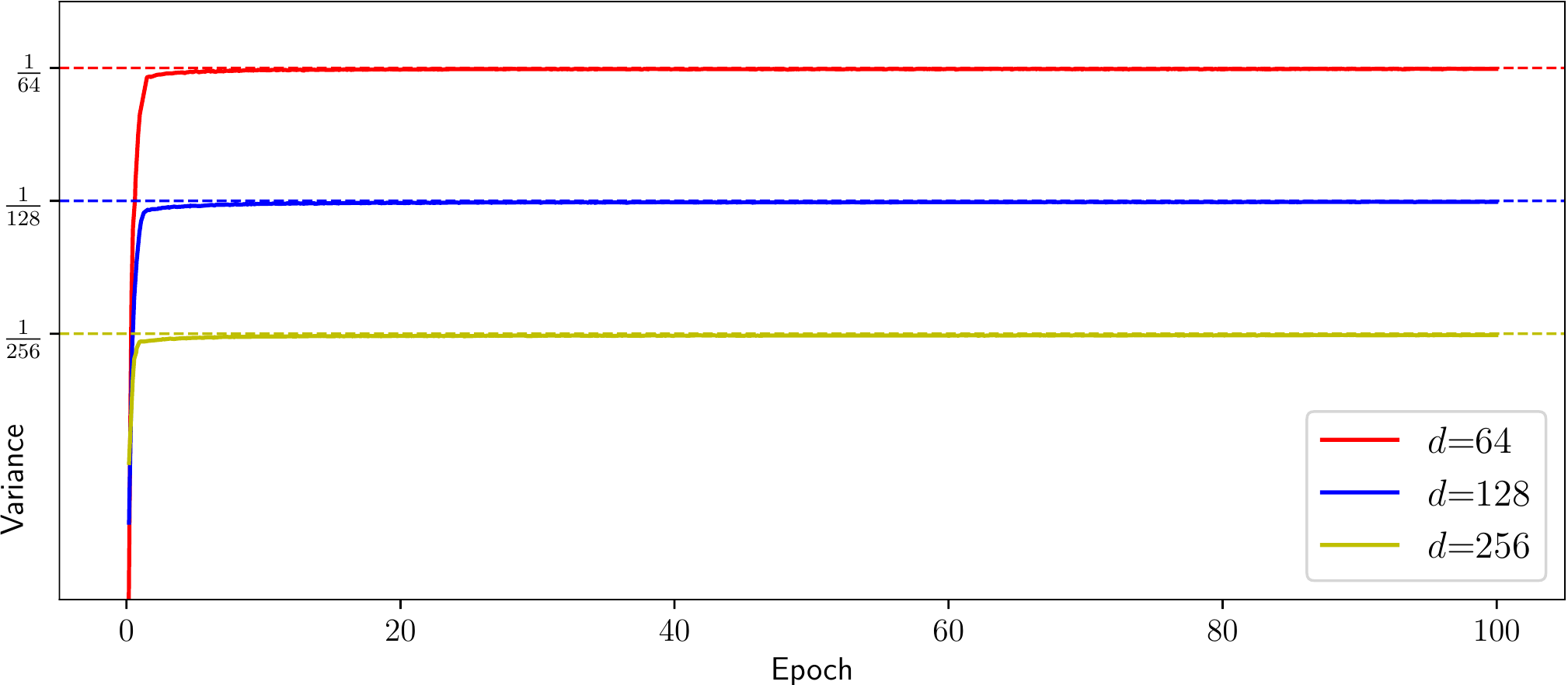}
  \vspace{-.5em}
  \caption{\textbf{Cross-image variance} tracked during the 100-epoch training process for our MoCo v2 baseline, with three encoding dimension options: $d{\in}\{64, 128, 256\}$. All of them quickly converge to $1/d$ (dotted lines). \label{fig:query_inter_var}}
  \vspace{-.5em}
\end{figure}

In this section, we show evidence with our MoCo v2 baseline that cross-image variance quickly converges to a constant that only depends on the encoding dimension $d$. This is through a monitor installed on the output encodings during training. Specifically, for each iteration, we compute the variance of the output $\ell_2$-normalized vectors from the source encoder along the \emph{batch} axis and average them over the \emph{channel} axis. Since each training batch contains different images rather than different views of the same image, the resulting value reflects the cross-image variance. Three encoding dimensions, $d{\in}\{64, 128, 256\}$ are experimented, and their variances during the 100-epoch training process are separately recorded in \cref{fig:query_inter_var}.

From the plot, we find that all the variances quickly and separately converge to $1/d$. For example, when the encoding dimension $d$ is $128$ (default), the variance converges to $1/128$; when $d$ is $64$, it converges to $1/64$. The same observations are made regardless of other designs for the encoder (\eg, \mcrop or \syncbn). We believe it is a natural outcome of Siamese representation learning which generally encourages \emph{uniformity}~\cite{wang2020understanding,chen2020exploring} -- encodings of different images distribute uniformly on the unit hypersphere. Therefore, cross-image variance is deemed \emph{not} an ideal reference to distinguish designs. Instead, we use intra-image variance which has a much smaller magnitude (\x$10^{-4}$), but carries useful signals to tell different designs apart (see \cref{fig:cases}). 

\section{ScaleMix\label{sec:scalemix_appendix}}
The goal of ScaleMix is to generate a new view $\vecv^{s}$ by combining two random sampled views of the same size (height $H$ and width $W$): $\vecv^1$ and $\vecv^2$. The generated new view is treated as a normal view of the input image $\vecx$ and used for Siamese learning. Specifically, following the protocol of~\cite{yun2019}, we define the combining operation as:
\begin{equation}
    \vecv^s = M \cdot \vecv^1 + (1 - M) \cdot \vecv^2, \nonumber
\end{equation}
where $M{\in}\{0,1\}^{H{\times}W}$ denotes a binary mask indicating where to use pixels from which view, and $\cdot$ is an element-wise multiplication. Note that different from other mixing operations~\cite{zhang2017mixup,yun2019}, we do not mix \emph{outputs} as both views are from the same image.

The binary values in $M$ are determined by bounding box coordinates $B{=}\left(x,y,w,h\right)$, where $(x,y)$ is the box center, and $(w,h)$ is the box size. Given $B$, its corresponding region in $M$ is set to all $0$ and otherwise all $1$. Intuitively, this means the region $B$ in $\vecv^1$ is removed and filled with the patch cropped from $B$ of $\vecv^2$.

The box coordinates $B$ are randomly sampled. We keep the aspect ratio of $B$ \emph{fixed} and the same as the input views, and only vary the size of the box according to a random variable $\lambda$ uniformly drawn from $(0,1)$: $w{=}W\sqrt{\lambda}$, $h{=}H\sqrt{\lambda}$. Box centers $(x,y)$ are again uniformly sampled.

\section{Detailed Theoretical Analysis\label{sec:theory_appendix}}

Given the outputs: $\vecz$ from the source encoder and $\vecz'$ from the target encoder (prime $'$ indicates target-related), the InfoNCE~\cite{oord2018representation} loss used by MoCo is defined as:
\begin{equation}
    \mathcal{L} := -\frac{1}{N}\sum_{i=1}^N \log \frac{\exp(\vecS_{ii'}/\tau)}{\epsilon \exp(\vecS_{ii'}/\tau) + \sum_{j\neq i} \exp(\vecS_{ij'}/\tau)}, \label{eq:loss_ap}
\end{equation}
where $N$ is batch size, $\tau$ is temperature, $\vecS_{ii'}{=}\vecz^\top_i \vecz'_i$ and $\vecS_{ij'}{=}\vecz^\top_i\vecz'_j$ are pairwise similarities between source and target encodings. We additionally introduce the parameter $\epsilon$ that controls the weight for the positive term in the denominator, where for standard loss $\epsilon{=}1$. 

For MoCo, only the source encoder receives gradient, and we take derivatives only for $\vecz_i$:
\begin{equation}
    \frac{\partial \mathcal{L}}{\partial \vecz_i} = \frac{1}{\tau}\sum_{j\neq i} \alpha_{ii'j'}(\vecz'_j - \vecz_i'), \label{eq:der_zi}
\end{equation}
where
\begin{equation}
    \alpha_{ii'j'} = \frac{\exp(\vecS_{ij'}/\tau-\vecS_{ii'}/\tau)}{\epsilon + \sum_{k\neq i} \exp(\vecS_{ik'}/\tau-\vecS_{ii'}/\tau)}. \label{eq:alpha_complete}
\end{equation}
For the simplified case where $\epsilon{=}0$~\cite{yeh2021decoupled}, we can have:
\begin{equation}
    \alpha_{ii'j'} = \alpha_{ij'} = \frac{\exp(\vecS_{ij'}/\tau)}{\sum_{k\neq i} \exp(\vecS_{ik'}/\tau)}, \label{eq:alpha_simple}
\end{equation}
which is independent of target encoding $\vecz_i'$.

Now, let's consider the last linear layer immediately before $\vecz$ as an example for analysis. Let $\vecf$ be the input features of this layer, $W$ be its weight matrix (so $\vecz{=}W\vecf$ and we do not consider $\ell_2$ normalization applied to $\vecz$). In this case, we can write down the dynamics of the source weight $W$ based on the gradient descent rule:
\begin{eqnarray}
    \dot W &:=& -\frac{\partial \mathcal{L}}{\partial W} = -\frac{1}{N}\sum_{i=1}^N \frac{\partial \mathcal{L}}{\partial \vecz_i} \vecf_i^\top \\
    &=& -\frac{1}{\tau N}\sum_{i=1}^N \sum_{j\neq i} \alpha_{ij'}(\vecz'_j - \vecz_i')\vecf_i^\top,
\end{eqnarray}
where $\dot W$ is a simplified notion of the change to \wrt $W$ following gradient decent.
Since both $\vecz'_j$ and $\vecz'_i$ come from the target encoder weight $W'$, we have $\vecz'_j{=}W' \vecf'_j$ and $\vecz_i'{=}W' \vecf'_i$ and thus:
\begin{equation}
    \dot W = -W' \frac{1}{\tau N}\sum_{i=1}^N \sum_{j\neq i} \alpha_{ij'}(\vecf'_j - \vecf_i')\vecf_i^\top \label{eq:gd-on-f}
\end{equation}

We define $\bar\vecf {:=} \ee[\vecf]$ to be the mean of the input feature and $\Sigma_\vecf {:=} \var[\vecf]$ to be the co-variance matrix of the input feature $\vecf$, where $\ee[\cdot]$ computes expectation and $\var[\cdot]$ outputs variance. These two quantities will be used later.

Now let's consider how intra-image variance in both target and source sides affect training. To reach a clear conclusion, we now make two assumptions.

\paragraph{Assumption 1: additive noise.} We can model the intra-image variance as \emph{additive noise}. Specifically, let $\tilde\vecf$ be the feature corresponding to the original image, we can assume:
\begin{itemize}
    \item $\vecf_i{=}\tilde \vecf_i {+} \vece_{i}$. That is, the input feature of the last layer $\vecf_i$ receives source noise $\vece_{i}$ with $\ee[\vece_{i}] {=} \bar \vece$ and $\var[\vece_{i}] {=} \Sigma$; 
    \item $\vecf'_j {=} \tilde \vecf_j {+} \vece'_{j}$. That is, the input feature $\vecf'_j$ receives target noise $\vece'_{j}$ with $\ee[\vece'_{j}] {=} \bar \vece'$ and $\var[\vece'_{j}] {=} \Sigma'$. Note that for the feature of a different image $\vecf'_{i}$, it also undergoes the same process on the target side and thus we have $\vecf'_{i} {=} \tilde \vecf_i {+} \vece'_{i}$. 
\end{itemize}
Note that the noise is not necessarily zero mean-ed. Since the augmentations of $\vecf_i$ and $\vecf'_{i}$ are independent, $\vece_{i}$ and $\vece'_{i}$ are independent of each other: $\pr(\vece_{i}, \vece'_{i}) {=} \pr(\vece_{i})\pr(\vece'_{i})$. Same for $\vece_{i}$ and $\vece_{j}$ where $i{\neq}j$.

\paragraph{Assumption 2: all $\alpha_{ij'}$ are constant and independent of $\vecf$.} Alternatively, if we consider the quadratic loss (\ie, $\mathcal{L}_q {=} \sum_{j\neq i} \left(\vecS_{ij'} {-} \vecS_{ii'}\right)$), then all $\alpha_{ij'}$ are constant and this assumption holds true. For InfoNCE this may not hold, and we leverage this assumption for simplicity of derivations. 

Under these two assumptions, we now compute $\ee_{\vecf}[\dot W]$, the \emph{expectation} of the weight gradient over \emph{input feature} $\vecf$ of the last layer. This gets rid of inter-image variance, and focuses on intra-image variance only:  
\begin{equation}
   \ee_{\vecf}[\dot W] = \frac{1}{\tau} W' (\Sigma_\vecf - R).
\end{equation}
Here the residual term $R$ is as follows: 
\begin{equation}
      R := -\frac{1}{N}\sum_{i=1}^N \hat\vece'_{i}(\bar \vecf + \vece_{i})^\top,
\end{equation}
where $\hat \vece'_{i} {:=} \sum_{j\neq i} \alpha_{ij'} \vece'_{j} {-} \vece'_{i}$ is also a random variable which is a weighted sum of $\vece'_{j}$ and $\vece'_{i}$. 

From the definition (\cref{eq:alpha_complete}), we have $\sum_{j\neq i} \alpha_{ij'} {=} 1$. $\vece'_{j}$ and $\vece'_{i}$ are independent. Therefore we can compute the mean and variance of $\hat \vece'_{i}$ as: 
\begin{eqnarray}
    \ee[\hat \vece'_{i}] &=& 0, \\
    \hat\Sigma'_{i} := \var[\hat\vece'_{i}] &=& (1 + \sum_{j\neq i}\alpha^2_{ij'})\Sigma'.
\end{eqnarray}

Now for the residual term $R$, we also have $\ee_{\vece}[R] {=} 0$. Therefore, the full expectation for $\dot W$ can be written as:
\begin{equation}
   \ee[\dot W] := \ee_{\vece}[\ee_{\vecf}[\dot W]] = \frac{1}{\tau} W' \Sigma_\vecf. \label{eq:w-theta-mean}
\end{equation}
This means the source weight will grow along the direction that maximizes the distance between different images. More precisely, it grows along the eigenvector that corresponds to the maximal eigenvalue of $\Sigma_\vecf$.

Now we can check the influence of \emph{intra-image} variance from source and target encoders. The influence can be characterized by the term $\var_{\vece}[\ee_{\vecf}[\dot W]]$. For simplicity, we can compute $\var_{\vece}[\ee_{\vecf}[\tr(R)]]$ -- \ie the variance on the trace of $R$, since $\Sigma_{\vecf}$ remains constant for intra-image variance.

Leveraging the independence of $\{\hat\vece'_{i}, \vece_{i}\}$ among different images, we can arrive at:
\begin{equation}
    \var_{\vece}[\ee_{\vecf}[\tr(R)]] = \tr\left[\hat\Sigma' (\bar\vecf \bar\vecf^\top + \bar\vece\bar\vece^\top + \Sigma)\right], \label{eq:trR-variance}
\end{equation}
where $\hat\Sigma' {:=} \frac{1}{N}\sum_{i=1}^N \hat\Sigma'_{i}$ is the mean of all variances of $\hat \vece'_{i}$.

From \cref{eq:trR-variance} we can notice that: i) if there is large magnitude of source feature mean $\bar\vecf$ and/or source noise mean $\bar \vece$, then the variance will be large; ii) this effect will be \emph{magnified} with more target-side variance (\ie, larger eigenvalues of $\Sigma'$ and thus $\hat \Sigma'$), \emph{without} affecting the average gradient; iii) large magnitude of feature mean and/or noise mean on the target side does \emph{not} influence the variance. This \emph{asymmetry} between source and target suggests that the training procedure an be negatively affected if the target variance is too large, coupled by $\bar\vecf \bar\vecf^\top$ and $\bar\vece\bar\vece^\top$ in \cref{eq:trR-variance}.

The intuition why there is such an asymmetry is the following: in~\cref{eq:gd-on-f}, while the target side has a subtraction $\vecf'_j - \vecf'_i$ which cancels out the mean, the source side $\vecf_i$ doesn't. This leads to the mean values being kept on the source side which couples with the target variance, whereas no mean values from the target side are kept.

Therefore, we can infer that higher variance on the target side is less necessary compared to the source side -- it will incur more instability during training without affecting the mean of gradients.

\section{More Implementation Details\label{sec:details_appendix}}

\paragraph{\mcrop.} Our \mcrop recipe largely follows the work of SwAV~\cite{caron2020unsupervised}. Specifically, 224-sized crops are sampled with a scale range of $(0.14,1)$, and 96-sized small crops are sampled from $(0.05,0.14)$. We use $m{=}6$ small crops by default, and each is forwarded separately with the encoder. When applied to one encoder, all $(1{+}6){=}7$ encodings are compared against the single encoding from the other side; when applied jointly, $(7{\times}2){=}14$ encodings are paired by crop size to compute loss terms. Unlike the practice in SwAV, no loss symmetrization is employed and the $6$ losses from small crops are averaged before adding to the standard loss. When target encoder is involved in \mcrop, we also create a separate memory bank~\cite{he2020momentum} dedicated to small crops, updated with $1$ out of the $6$ crops.

\paragraph{\aaug.} For \saug, we use additional augmentations from RandAug~\cite{cubuk2020randaugment}, same as~\cite{wang2021contrastive}. For \waug, we simply remove all the color- and blur-related augmentations and only keep geometric ones in the MoCo v2 recipe. This leaves us with random resized cropping and flipping.

\paragraph{\menc.} Deviating from \mcrop, augmentations used for computing the mean are forwarded \emph{jointly} through the encoder thanks to the uniform size of 224\x224. Joint forwarding enlarges the batch size in BN, which further reduces the variance. The output encodings are averaged before $\ell_2$ normalization. 

\paragraph{Other frameworks.} Different from MoCo v2 which uses shuffle BN~\cite{he2020momentum} across 8 GPUs, all the frameworks studied in \cref{sec:frameworks} use \syncbn by default. Therefore, when applying \abn to them, we keep the target encoder untouched and change the BNs in the source encoder instead. To \emph{minimize} the impact from the number of GPU devices (\eg, MoCo v3 uses 16 GPUs to fit a batch size of 4096 for ResNet; whereas for ViT it uses 32 GPUs), we always divide the full batch into 8 \emph{groups} and the normalization is performed within each group -- this mimics the per-device BN operation in MoCo v2 while being more general.

Moreover, for MoCo v2 we only convert the single BN in the target projector to \syncbn. This has minimal influence on efficiency as \syncbn can be expensive and converting all of them (including ones in the encoder) can significantly slow down training. Now since we are converting \syncbn \emph{back}, we choose to convert \emph{all} BNs in the source encoder whenever possible to reduce inter-device communications for efficiency purposes.

More recent frameworks~\cite{chen2021empirical,zbontar2021barlow} adopt the asymmetric augmentation recipe in BYOL~\cite{grill2020bootstrap}, in such cases, we use one composition for source and the other for target half the time during pre-training, and swap them in the other half.

To have a fair comparison with frameworks pre-trained for 100 epochs, we optionally train 2\x{} as long when the default loss is symmetrized and ours is asymmetric.

Unless otherwise specified, we follow the same design choices in MoCo v2 when applying \smix and \menc to other frameworks. In addition, there are subtleties associated with each individual framework listed below:
\begin{itemize}
    \item \textbf{MoCo v3}~\cite{chen2021empirical}. Since MoCo v3 also employs an additional predictor on the source side, we involve both the predictor and the backbone when applying \abn. 
    \item \textbf{SimCLR}~\cite{chen2020simple}. The original SimCLR uses $2{\times}N{-}2$ negative examples for contrastive learning~\cite{chen2020simple}, which includes all the other images in the same batch, multiplied by 2 for the two augmentations per image. After converting to the asymmetric version, we only use $N{-}1$ negative samples -- same as in MoCo v3 -- and it causes a gap. We find a simple \emph{change} of InfoNCE~\cite{oord2018representation} temperature from $0.1$ to $0.2$ can roughly compensate for this gap. For \abn, we convert all the BNs in the source encoder, not just the ones in the projector. For \smix, we apply this augmentation half the time -- we empirically find applying \smix all the time will cause a considerable drop in performance compared to the asymmetric baseline, for reasons yet to be understood.
    \item \textbf{BYOL}~\cite{grill2020bootstrap}. BYOL initiated the additional predictor which also has BNs. We convert all the BNs in the source encoder when \abn is used, not just ones in the projector.
    \item \textbf{SimSiam}~\cite{chen2020exploring}. Additional predictor is again used in SimSiam and plays an important role in collapse prevention. We convert all the BNs in the source encoder after the conversion to an asymmetric version.
    \item \textbf{Barlow Twins}~\cite{zbontar2021barlow}. This is a fully symmetric framework and \emph{no} loss symmetrization is used by default. Therefore, we also pre-train the asymmetric version for 100 epochs, not 2\x{} as long. Same as SimCLR, \smix is applied with half the frequency. All the encoder BNs are converted when \abn is used.
\end{itemize}

\paragraph{ViT backbone.} MoCo v3~\cite{chen2021empirical} with its default hyper-parameters for ViT backbone is used. ViT as a backbone does \emph{not} have BN. Therefore we convert BNs in the projector and predictor when using \abn.

\paragraph{Transfer learning.} We follow the linear probing protocol to evaluate our model on transfer learning tasks. Different from ImageNet, we use SGD optimizer with momentum $0.9$ and weight decay $0$ for training. The learning rate is adjusted via grid search on the validation set, and the final results are reported on the test set. All models are trained for 100 epochs, with a half-cycle cosine decaying schedule for learning rate.

{\small
\bibliographystyle{ieee_fullname}
\bibliography{local}

\begin{thebibliography}{10}\itemsep=-1pt

\bibitem{bardes2021vicreg}
Adrien Bardes, Jean Ponce, and Yann LeCun.
\newblock Vicreg: Variance-invariance-covariance regularization for
  self-supervised learning.
\newblock {\em arXiv preprint arXiv:2105.04906}, 2021.

\bibitem{berthelot2019mixmatch}
David Berthelot, Nicholas Carlini, Ian Goodfellow, Nicolas Papernot, Avital
  Oliver, and Colin~A Raffel.
\newblock Mixmatch: A holistic approach to semi-supervised learning.
\newblock In {\em NeurIPS}, 2019.

\bibitem{bertinetto2016fully}
Luca Bertinetto, Jack Valmadre, Joao~F Henriques, Andrea Vedaldi, and Philip~HS
  Torr.
\newblock Fully-convolutional siamese networks for object tracking.
\newblock In {\em ECCV}, 2016.

\bibitem{Bromley1994}
Jane Bromley, Isabelle Guyon, Yann LeCun, Eduard S{\"a}ckinger, and Roopak
  Shah.
\newblock Signature verification using a {``Siamese"} time delay neural
  network.
\newblock In {\em NeurIPS}, 1994.

\bibitem{cai2021exponential}
Zhaowei Cai, Avinash Ravichandran, Subhransu Maji, Charless Fowlkes, Zhuowen
  Tu, and Stefano Soatto.
\newblock Exponential moving average normalization for self-supervised and
  semi-supervised learning.
\newblock In {\em CVPR}, 2021.

\bibitem{caron2020unsupervised}
Mathilde Caron, Ishan Misra, Julien Mairal, Priya Goyal, Piotr Bojanowski, and
  Armand Joulin.
\newblock Unsupervised learning of visual features by contrasting cluster
  assignments.
\newblock {\em arXiv preprint arXiv:2006.09882}, 2020.

\bibitem{caron2021emerging}
Mathilde Caron, Hugo Touvron, Ishan Misra, Herv{\'e} J{\'e}gou, Julien Mairal,
  Piotr Bojanowski, and Armand Joulin.
\newblock Emerging properties in self-supervised vision transformers.
\newblock {\em arXiv preprint arXiv:2104.14294}, 2021.

\bibitem{chen2020simple}
Ting Chen, Simon Kornblith, Mohammad Norouzi, and Geoffrey Hinton.
\newblock A simple framework for contrastive learning of visual
  representations.
\newblock {\em arXiv preprint arXiv:2002.05709}, 2020.

\bibitem{chen2020improved}
Xinlei Chen, Haoqi Fan, Ross Girshick, and Kaiming He.
\newblock Improved baselines with momentum contrastive learning.
\newblock {\em arXiv preprint arXiv:2003.04297}, 2020.

\bibitem{chen2020exploring}
Xinlei Chen and Kaiming He.
\newblock Exploring simple siamese representation learning.
\newblock In {\em CVPR}, 2021.

\bibitem{chen2021empirical}
Xinlei Chen, Saining Xie, and Kaiming He.
\newblock An empirical study of training self-supervised vision transformers.
\newblock {\em arXiv preprint arXiv:2104.02057}, 2021.

\bibitem{cubuk2020randaugment}
Ekin~D Cubuk, Barret Zoph, Jonathon Shlens, and Quoc~V Le.
\newblock Randaugment: Practical automated data augmentation with a reduced
  search space.
\newblock In {\em CVPRW}, 2020.

\bibitem{deng2009imagenet}
Jia Deng, Wei Dong, Richard Socher, Li-Jia Li, Kai Li, and Li Fei-Fei.
\newblock Imagenet: A large-scale hierarchical image database.
\newblock In {\em CVPR}, 2009.

\bibitem{Dosovitskiy2021}
Alexey Dosovitskiy, Lucas Beyer, Alexander Kolesnikov, Dirk Weissenborn,
  Xiaohua Zhai, Thomas Unterthiner, Mostafa Dehghani, Matthias Minderer, Georg
  Heigold, Sylvain Gelly, Jakob Uszkoreit, and Neil Houlsby.
\newblock An image is worth 16x16 words: Transformers for image recognition at
  scale.
\newblock In {\em ICLR}, 2021.

\bibitem{dwibedi2021little}
Debidatta Dwibedi, Yusuf Aytar, Jonathan Tompson, Pierre Sermanet, and Andrew
  Zisserman.
\newblock With a little help from my friends: Nearest-neighbor contrastive
  learning of visual representations.
\newblock {\em arXiv preprint arXiv:2104.14548}, 2021.

\bibitem{everingham2010pascal}
Mark Everingham, Luc Van~Gool, Christopher~KI Williams, John Winn, and Andrew
  Zisserman.
\newblock The pascal visual object classes (voc) challenge.
\newblock {\em IJCV}, 2010.

\bibitem{Goyal2017}
Priya Goyal, Piotr Doll{\'a}r, Ross Girshick, Pieter Noordhuis, Lukasz
  Wesolowski, Aapo Kyrola, Andrew Tulloch, Yangqing Jia, and Kaiming He.
\newblock Accurate, large minibatch {SGD}: Training {ImageNet} in 1 hour.
\newblock {\em arXiv:1706.02677}, 2017.

\bibitem{grill2020bootstrap}
Jean-Bastien Grill, Florian Strub, Florent Altch{\'e}, Corentin Tallec,
  Pierre~H Richemond, Elena Buchatskaya, Carl Doersch, Bernardo~Avila Pires,
  Zhaohan~Daniel Guo, Mohammad~Gheshlaghi Azar, et~al.
\newblock Bootstrap your own latent: A new approach to self-supervised
  learning.
\newblock {\em arXiv preprint arXiv:2006.07733}, 2020.

\bibitem{he2020momentum}
Kaiming He, Haoqi Fan, Yuxin Wu, Saining Xie, and Ross Girshick.
\newblock Momentum contrast for unsupervised visual representation learning.
\newblock In {\em CVPR}, 2020.

\bibitem{he2016deep}
Kaiming He, Xiangyu Zhang, Shaoqing Ren, and Jian Sun.
\newblock Deep residual learning for image recognition.
\newblock In {\em CVPR}, 2016.

\bibitem{ioffe2015batch}
Sergey Ioffe and Christian Szegedy.
\newblock Batch normalization: Accelerating deep network training by reducing
  internal covariate shift.
\newblock In {\em ICML}, 2015.

\bibitem{jiang2021all}
Zihang Jiang, Qibin Hou, Li Yuan, Daquan Zhou, Yujun Shi, Xiaojie Jin, Anran
  Wang, and Jiashi Feng.
\newblock All tokens matter: Token labeling for training better vision
  transformers.
\newblock {\em arXiv preprint arXiv:2104.10858}, 2021.

\bibitem{Krizhevsky2009}
Alex Krizhevsky.
\newblock Learning multiple layers of features from tiny images.
\newblock {\em Tech Report}, 2009.

\bibitem{li2021momentum}
Zeming Li, Songtao Liu, and Jian Sun.
\newblock Momentum$^{2}$ teacher: Momentum teacher with momentum statistics for
  self-supervised learning.
\newblock {\em arXiv preprint arXiv:2101.07525}, 2021.

\bibitem{loshchilov2016sgdr}
Ilya Loshchilov and Frank Hutter.
\newblock Sgdr: Stochastic gradient descent with warm restarts.
\newblock {\em arXiv preprint arXiv:1608.03983}, 2016.

\bibitem{maji2013fine}
Subhransu Maji, Esa Rahtu, Juho Kannala, Matthew Blaschko, and Andrea Vedaldi.
\newblock Fine-grained visual classification of aircraft.
\newblock {\em arXiv preprint arXiv:1306.5151}, 2013.

\bibitem{misra2020self}
Ishan Misra and Laurens van~der Maaten.
\newblock Self-supervised learning of pretext-invariant representations.
\newblock In {\em CVPR}, 2020.

\bibitem{oord2018representation}
Aaron van~den Oord, Yazhe Li, and Oriol Vinyals.
\newblock Representation learning with contrastive predictive coding.
\newblock {\em arXiv preprint arXiv:1807.03748}, 2018.

\bibitem{yun2019}
Yun Sangdoo, Han Dongyoon, Oh Seong, Joon, Chun Sanghyuk, Choe Junsuk, and Yoo
  Youngjoon.
\newblock Cutmix: Regularization strategy to train strong classifiers with
  localizable features.
\newblock In {\em ICCV}, 2019.

\bibitem{sohn2020fixmatch}
Kihyuk Sohn, David Berthelot, Chun-Liang Li, Zizhao Zhang, Nicholas Carlini,
  Ekin~D Cubuk, Alex Kurakin, Han Zhang, and Colin Raffel.
\newblock Fixmatch: Simplifying semi-supervised learning with consistency and
  confidence.
\newblock {\em arXiv preprint arXiv:2001.07685}, 2020.

\bibitem{taigman2014deepface}
Yaniv Taigman, Ming Yang, Marc'Aurelio Ranzato, and Lior Wolf.
\newblock Deepface: Closing the gap to human-level performance in face
  verification.
\newblock In {\em CVPR}, 2014.

\bibitem{tarvainen2017mean}
Antti Tarvainen and Harri Valpola.
\newblock Mean teachers are better role models: Weight-averaged consistency
  targets improve semi-supervised deep learning results.
\newblock In {\em NeurIPS}, 2017.

\bibitem{tian2021understanding}
Yuandong Tian, Xinlei Chen, and Surya Ganguli.
\newblock Understanding self-supervised learning dynamics without contrastive
  pairs.
\newblock {\em arXiv preprint arXiv:2102.06810}, 2021.

\bibitem{tian2020understanding}
Yuandong Tian, Lantao Yu, Xinlei Chen, and Surya Ganguli.
\newblock Understanding self-supervised learning with dual deep networks.
\newblock {\em arXiv preprint arXiv:2010.00578}, 2020.

\bibitem{wang2020understanding}
Tongzhou Wang and Phillip Isola.
\newblock Understanding contrastive representation learning through alignment
  and uniformity on the hypersphere.
\newblock In {\em ICML}, 2020.

\bibitem{wang2019enaet}
Xiao Wang, Daisuke Kihara, Jiebo Luo, and Guo-Jun Qi.
\newblock Enaet: Self-trained ensemble autoencoding transformations for
  semi-supervised learning.
\newblock {\em arXiv preprint arXiv:1911.09265}, 2019.

\bibitem{wang2021contrastive}
Xiao Wang and Guo-Jun Qi.
\newblock Contrastive learning with stronger augmentations.
\newblock {\em arXiv preprint arXiv:2104.07713}, 2021.

\bibitem{wang2018iterative}
Yisen Wang, Weiyang Liu, Xingjun Ma, James Bailey, Hongyuan Zha, Le Song, and
  Shu-Tao Xia.
\newblock Iterative learning with open-set noisy labels.
\newblock In {\em CVPR}, 2018.

\bibitem{wu2018group}
Yuxin Wu and Kaiming He.
\newblock Group normalization.
\newblock In {\em ECCV}, 2018.

\bibitem{wu2018unsupervised}
Zhirong Wu, Yuanjun Xiong, Stella~X Yu, and Dahua Lin.
\newblock Unsupervised feature learning via non-parametric instance
  discrimination.
\newblock In {\em CVPR}, 2018.

\bibitem{xu2020seed}
Haohang Xu, Xiaopeng Zhang, Hao Li, Lingxi Xie, Hongkai Xiong, and Qi Tian.
\newblock Seed the views: Hierarchical semantic alignment for contrastive
  representation learning.
\newblock {\em arXiv preprint arXiv:2012.02733}, 2020.

\bibitem{yeh2021decoupled}
Chun-Hsiao Yeh, Cheng-Yao Hong, Yen-Chi Hsu, Tyng-Luh Liu, Yubei Chen, and Yann
  LeCun.
\newblock Decoupled contrastive learning.
\newblock {\em arXiv preprint arXiv:2110.06848}, 2021.

\bibitem{You2017}
Yang You, Igor Gitman, and Boris Ginsburg.
\newblock Large batch training of convolutional networks.
\newblock {\em arXiv:1708.03888}, 2017.

\bibitem{zbontar2021barlow}
Jure Zbontar, Li Jing, Ishan Misra, Yann LeCun, and St{\'e}phane Deny.
\newblock Barlow twins: Self-supervised learning via redundancy reduction.
\newblock {\em arXiv preprint arXiv:2103.03230}, 2021.

\bibitem{zhang2017mixup}
Hongyi Zhang, Moustapha Cisse, Yann~N. Dauphin, and David Lopez-Paz.
\newblock mixup: Beyond empirical risk minimization.
\newblock {\em arXiv preprint arXiv:1710.09412}, 2017.

\bibitem{zhou2014learning}
Bolei Zhou, Agata Lapedriza, Jianxiong Xiao, Antonio Torralba, and Aude Oliva.
\newblock Learning deep features for scene recognition using places database.
\newblock In {\em NeurIPS}, 2014.

\bibitem{zhou2021theory}
Pan Zhou, Caiming Xiong, Xiao-Tong Yuan, and Steven Hoi.
\newblock A theory-driven self-labeling refinement method for contrastive
  representation learning.
\newblock {\em arXiv preprint arXiv:2106.14749}, 2021.

\end{thebibliography}
}

\end{document}